\documentclass[journal]{IEEEtran}
\ifCLASSINFOpdf
\else
\fi
\usepackage{mathtools}
\DeclarePairedDelimiter\abs{\lvert}{\rvert}%

\newcommand{\argmax}{\arg\!\max}

\usepackage[english]{babel}

\usepackage{cite} 

\usepackage{times}
\usepackage{amssymb}
\usepackage{soul}
\usepackage{url}
\usepackage[utf8]{inputenc}
\usepackage[small,center]{caption}
\usepackage{booktabs}
\usepackage{algorithmic}
\usepackage{subfig}
\usepackage{ragged2e}
\usepackage{float}
\usepackage{amsmath}
\usepackage{comment}
\usepackage{balance} 
\usepackage{graphicx} %package to manage images
\usepackage{wrapfig}
\usepackage[ruled,vlined ]{algorithm2e}
\graphicspath{ {Images/} }
\usepackage{multirow}
\begin{document}
\title{CRL: Class Representative Learning for Image Classification}

\author{Mayanka~ Chandrashekar,~\IEEEmembership{Member,~IEEE,}
       and Yugyung~Lee,~\IEEEmembership{Member,~IEEE}
        %and~Jane~Doe,~\IEEEmembership{Life~Fellow,~IEEE}% <-this % stops a space
\thanks{M.Chandrashekar (mckw9@mail.umkc.edu) and Y. Lee (leeyu@umkc.edu) are with the Department
of Computer Science and Electrical Engineering, University of Missouri, Kansas City,
MO, 64110 USA}% <-this % stops a space
%\thanks{Manuscript received June 30, 2017; revised July 26, 2017.}
}

\maketitle

\begin{abstract}
Building robust and real-time classifiers with diverse datasets are one of
the most significant challenges to deep learning researchers.
It is because there is a considerable gap between a model
built with training (seen) data and  real (unseen) data in applications.
Recent works including Zero-Shot Learning  (ZSL), have attempted to deal 
with this problem of overcoming the apparent gap through transfer learning.
In this paper, we propose a novel model, called Class Representative Learning Model (CRL), 
that  can be especially effective in image classification influenced by ZSL.
In the CRL model, first, the learning step is to build class representatives to represent classes in datasets
by aggregating prominent features extracted from a  Convolutional Neural Network (CNN). 
Second, the inferencing step in CRL is to match between the class representatives and new data. 
The proposed CRL model demonstrated superior performance compared to the current state-of-the-art research in ZSL and mobile deep learning. 
The proposed CRL model has been implemented and evaluated in a parallel environment, using Apache Spark, for both distributed learning and recognition. An extensive experimental study on  the benchmark datasets, ImageNet-1K, CalTech-101, CalTech-256, CIFAR-100, shows that CRL can build a class distribution model with drastic improvement in learning and recognition performance without sacrificing accuracy compared to the state-of-the-art performances in image classification.
\end{abstract}

% \begin{IEEEkeywords}
% IEEE, IEEEtran, journal, \LaTeX, paper, template.
% \end{IEEEkeywords}

% \fi
\IEEEpeerreviewmaketitle

\section{Introduction}

%%0) Deep Visual Embeddings & K-means

Recent advances in deep learning (DL) have improved
the state-of-the-art researches in the data-driven approaches and applications in a wide range of domains. However, building
robust and real-time classifiers with diverse datasets is one of the most significant challenges to deep learning researchers. 
It is because there is a considerable gap between a model built
with training (seen) data and  real (unseen) data in applications \cite{wang2019survey,bendale2015towards,lampert2009learning}.
The current deep learning research assumes strong boundaries between data, between data and models, 
and between models in deep learning.  
%There is no attempt made to  deal with this problem of breaking the boundaries or dynamically building a model. 
The new paradigm focuses on the universal representation of diverse datasets. % and dynamic modeling depending upon users’ context appear of great importance.
%%% 1) zero shot/few shot

There has been increasing attention on Zero-Shot Learning (ZSL) \cite{wang2019survey}  and one-shot/few-shot learning(FSL) \cite{lake2011one,vinyals2016matching}.
These efforts aim to build the ability to learn from a few examples or even without seeing them. Alternatively, it is required to represent and match new instances on a semantic space, which results in minimizing training efforts and maximizing learning outcomes. They focus on active transfer learning by fully leveraging information from pre-trained models. The seamless integration of unlabeled data from the seen/unseen classes is possible through the expressive representations of multi-model embeddings, including semantic, word, visual embeddings. However, there
are the notable limitations of the ZSL or FSL approaches: Many of them are relying on semantic embeddings in a common semantic space having a generative model \cite{kingma2014semi,diederik2014auto,rezende2016one}.

The recent ZSL works  demonstrated their effectiveness in transferred from prior experiences to new classes, which is a form of transfer learning. The most used semantic space in the ZSL model is supported by a joint embedding framework called Label-Embedding Space \cite{wang2019survey,xu2019complementary}  containing a combination of visual embeddings and word embeddings; or Engineering Semantic Space called
Attribute Space, which uses attribute annotations for the ZSL model \cite{akata2016multi}. In contrast to prior work, we mainly extract the deep neural network features learned from visual inputs of seen classes creating image representatives, and we do not rely on any other features such as attribute annotations or word embeddings.

Similar to our approach in the feature extraction, there are
active efforts \cite{mahendran2015understanding,zhou2016learning} for extracting important features from
Convolutional Neural Networks (CNNs) such as Inception or ResNet.
Mahendran et al. \cite{mahendran2015understanding} analyzed the preserved deep features
through inverting the fully-connected layers. Zhou et al. \cite{zhou2016learning} built the class activation map using CNN features for the
localization of the objects in the images for the discriminative
image regions. Unlike \cite{mahendran2015understanding} and \cite{zhou2016learning}, we are interested in
generating class representatives using CNN features.

The  goal  is  to  propose  an  innovative  model  called {\it Class Representative Learning (CRL)}  for  image  classification  for seen and unseen data. In this model, the focus is on creating a universal representation called the class representatives using the  source  environment,  which  is  typically  pre-trained  deep learning  models.  Given  this  goal,  architectural  improvements are  not  our  purpose;  instead,  we explore  universal representatives that could be used for classification.  It  is  desired to  enable  the  universal  representation  to  be  trained  from  any existing  architectures  or  datasets  with  reduced  efforts  and resources.  The  minimum  requirement  for  the  CRL  model  is to have a  suitable {\it source}  (pre-trained)  CNN model  that can be mapped to the given datasets ({\it target}).

The CRL model can be classified as transfer learning, called
meta-learning \cite{ravi2016optimization,yu2019meta}. The basic idea behind transfer
learning is to use previously learned knowledge on different domains or tasks. The CRL model is based on the transductive
approach that aims to project the target data onto a source
environment for the extraction of features by mapping to unify
the input spaces. The transductive property in transfer learning
is to derive the values of the unknown function for points of
interest (class-based or instance-based) from the given data
(source environment or source domain) \cite{wang2019survey,vapnik2013nature}.

The CRL model poses the property of being selective during inferencing. In other words, the CRL model can classify an input
image to either source labels or target labels or both. Due to
this property, the CRL model can behave like a traditional
classification model. The Convolution Neural Network-based Classification models tend to be high in parameter
requirements to achieve state-of-the-art accuracy \cite{canziani2016analysis,tan2019efficientnet}.
To show the superiority of the CRL model developed in this study,
we have compared our CRL model against other state-of-the-art deep learning models.

%%% Our contribution
The contributions of this paper can be summarized as
follows:
\begin{itemize}
\item The proposed model (CRL) is an efficient way of building class-level classifiers by utilizing features from a pre-trained model in Convolutional Neural Network for classification problems.

\item The CRL  has the ability to build models for the similarity distribution of CRs for given datasets and estimate the CR accuracies in the classification.
\item A comprehensive evaluation of the proposed model has been conducted with 
deep learning models, Zero-Shot Learning (ZSL) using the four benchmark datasets in terms of time and accuracy. 
 In addition,  CRL was compared with the MobileNet models  \cite{sandler2018mobilenetv2,sandler2018mobilenetv2,zoph2018learning}.
 \end{itemize}

\section{Related Work}

%https://machinelearningmastery.com/transduction-in-machine-learning/

\subsection{Transfer Learning}

Recent studies have indicated the importance of transfer learning (TL) \cite{tzeng2015simultaneous,kornblith2018better} that aims to maximize the learning outcome by transferring a model developed for a task for building a model on another task. NASNet \cite{zoph2018learning} explored the possibility of transferring from what learned from a small dataset (e.g., CIFAR-10) to a larger dataset (ImageNet-1K) through searching and utilizing a core architectural building block from the small dataset.

He et al. \cite{he2018rethinking} have shown that pre-trained models with an extensive data set like ImageNet-1K or with a small dataset like a subset of MS COCO have incredible influence in computer vision. Initialization with pre-trained models or evaluating with pre-trained features (e.g., unsupervised learning \cite{pathak2017learning}) can reduce efforts and produce better results in Deep Learning (DL). It is possible because pre-training models are widely available, and learning from the models is faster than building from scratch.

 The DL community has extensively studied transfer Learning \cite{tzeng2015simultaneous,kornblith2018better}. 
 The transfer learning from ImageNet-1K in Decaf \cite{donahue2014decaf}  showed substantial improvements compared to
learning from image features. Ravi et al. \cite{ravi2016optimization} also presented
a meta-learner model that supported the quick convergence of training with a new task using few-shot learning.
 
Pan et al. \cite{pan2010survey} defined an inductive transfer learning as
cross-domain learning where the target task is different from
the source task. The data in the target domain are required to
induce a predictive model that can be transferred from the
source domain to the target domain. Our model is similar
to the Feature-representation-transfer defined by Pan et al.
However, the difference is that our model was encoded based
on the aggregation of the high-level features extracted from
Convolutional Neural Networks.

In our paper, we used a pre-trained model only for feature
extraction, but training is not required with new data. After the
fully connected layers are removed from the entire network,
the rest will be mainly used for feature extraction for new
data. Thus, the use of the pre-trained model in our study is
different from others since we only use it as a reference model
for extracting features for new data.

\subsection{Universal Representation}

Ubernet \cite{kokkinos2017ubernet} is a {\it universal}  CNN that allows solving multiple tasks in a unified architecture efficiently. It is through
the end-to-end network training with a single training set for diverse datasets and low memory complexity. Universal
representations \cite{rebuffi2017learning,bilen2017universal} perform well for visual domains in a uniform manner and have proven to be efficient for multiple domain learning in relatively small neural networks.

Rebuff et al. \cite{rebuffi2018efficient} demonstrated that universal parametric
families of networks could share parameters among multiple
domains using parallel residual adapter modules. Similar to
our work, all these works presented universal representations
for multiple domains or multiple tasks. However,  unlike CRL, none of
them focus on dynamically generating a model for multiple
domains.

In this paper, we define {\it Source Environment} for providing a basis for feature selection as well as 
a uniform representation of a set of heterogeneous data sources for effective deep learning.
Feature selection is a crucial step in machine learning since it directly influences the performance of machine learning. (e.g., as the right choice of features drives the classifier to perform well).  However, Kapoor et al. \cite{kapoor2012learning} observed that finding useful features for multi-class classification is not trivial due to the volume in the high-dimensional feature space as well as the sparseness over the search space.

Dictionary learning \cite{shen2015multi} was presented to determine the subspaces and build dictionaries by efficiently reducing dimensionality for efficient representations of classes of images. The critical contribution of the work is the reduction of sparsity constraints and the improvement of accuracy by identification of the most essential components of the observed data. 

From the extracted set of relevant features from images and quantizing them with these bags of visual words, we will further build up a visual CR vector for each class by combining these primitive features. The visual CRs will be used for efficient learning as well as recognition with large scale multi-class datasets.

\subsection{Lightweight Deep Learning}

Recently, there has been an increasing demand for mobile applications for small networks or dynamic networks in deep learning. There have been several deep neural architectures to strike an optimal balance between accuracy and performance. The lightweight deep learning was achieved using three types of layer compression techniques namely: weight compression, Convolution compression and adding a single layer \cite{nan2019deep}. Weight compression is the primitive technique used to create a lightweight model.  MobileNet-V1 \cite{howard2017mobilenets} and Shufflenet \cite{zhang2018shufflenet} used a convolution compression technique in its architecture, specifically depth-wise separable convolutions and point-wise group convolution, respectively.

As an extension of the previous works, MobileNet-v2 \cite{sandler2018mobilenetv2} added a new layer, namely an inverted residual layer, with a narrow bottleneck to create a lightweight model. In NasNet Mobile \cite{zoph2018learning}, a new paradigm, called Neural Architecture Search (NAS), was proposed with reinforcement learning for knowledge transfer. In general, architectural changes are typically considered to achieve a lightweight model. In the CRL model, we have obtained a lightweight model through
the flexibility of representation concerning the class.

\subsection{Matching Networks}

Few-shot classification \cite{lake2011one,vinyals2016matching}  is to label new classes which
are not seen in training, but through matching with only a few examples of each of these classes. The matching networks
are similar to a weighted nearest-neighbor classifier in an embedding space. The embedding in the matching networks
was built as a form of sampled mini-batches, called episodes, during training. Notably, matching networks \cite{vinyals2016matching} is similar to
our work in terms of mapping an attention-based embedding to a query set for predicting classes. However, our model is
different from these works since we can build a dynamic model for multiple classes by assembling a set of a single class classifier, called Class Representative, 
built one by one independently.

A meta-learning approach \cite{ravi2016optimization}  aims to build a custom
model for each episode based on Long Short-Term Memory (LSTM), unlike others building
each episode over multiple episodes. The prototypical networks \cite{snell2017prototypical}  built a class prototype by computing the mean
of the training set in the embedding space and find the nearest class prototype for a query set as inferencing. This approach is
very similar to our work in terms of building the class’s prototype as an abstraction of the class by learning an embedding of the
meta-data into a shared space. Recently, Wang et al. \cite{wang2018zero} extended the performance of Zero-Shot Learning and 
Few-Shot Learning using latent-space distributions of discriminative feature representations. Similar to our approach, 
they used only the feature extractor of the CNN model. However, they are
different from our work in terms of the following aspects: (1) they used variational autoencoder (VAE) while we are using a vector space model with a cosine similarity measurement, (2) they built a model for all classes in any given dataset while we are building a model class by class. They focused on learning
an embedding of the meta-data into a shared space. However, in our work, we build CRs class by class. Thus, once the CR
is generated, there is no dependence between CRs. Due to the independence between CRs, we can build multi-class models dynamically.

\subsection{Zero-Shot Learning}

Zero-Shot Learning (ZSL) defined a semantic encoding for predicting new classes by using a standard feature set derived from a semantic knowledge base \cite{palatucci2009zero}. All well-known works have worked on learning and understanding explicit and external attributes. Much of the early work adapted from the original definition of the semantic knowledge base in Zero-Shot Learning \cite{palatucci2009zero}, focused on attributes solely based on visual feature learning. Some of the works in the feature learning include boosting techniques \cite{wolf2005robust}, object detection \cite{torralba2007sharing}, chopping algorithm \cite{fleuret2006pattern}, feature adaption \cite{bart2005cross}, and linear classifiers \cite{amit2007uncovering}. 

\begin{figure}
    \centering
    \includegraphics[width=\linewidth]{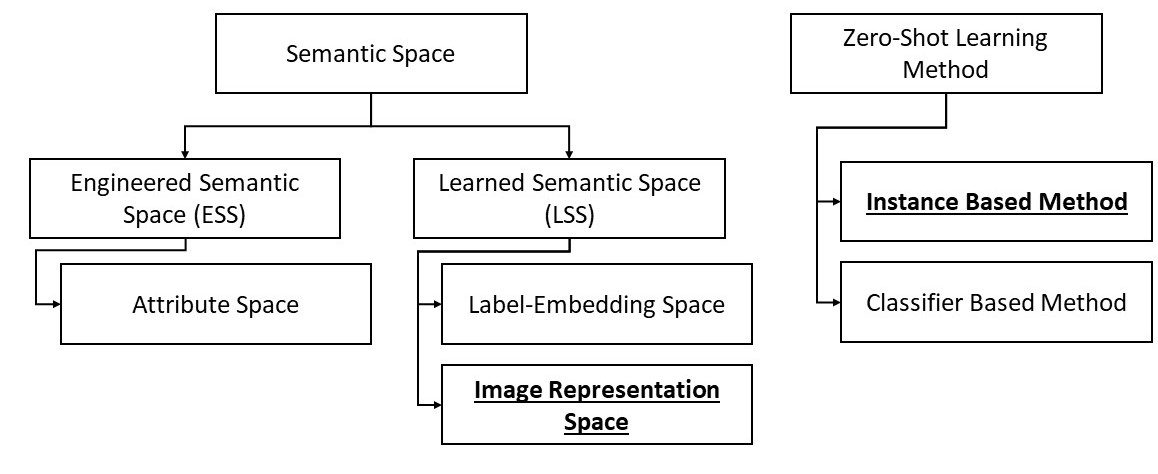}
    \caption{Types of Semantic Space and ZSL Methods}
    \label{fig:RelatedCRLZSL}
    \vspace{1mm}
    
    \footnotesize{* The CRL model belongs to the highlighted types.}
\end{figure}

The recent works of Zero-Shot Learning can be categorized into Engineered Semantic Spaces (ESS) and Learned
Semantic Spaces (LSS) according to the semantic space type
and ZSL methods in Wang et al. \cite{wang2019survey}.  ESS can be further sub-categorized into Attribute Space, Lexical Space, and
Text-Keyword Space; and LSS into Label-Embedding Space,
Text-Embedding Space, and Image-Representation Space. The Zero-Shot Learning method is classified into Classifier-based
Methods and Instance-Based Methods. According to this
categorization, the CRL model can be classified as Image
Representation Semantic Space and Instance-Based Method,
specifically Projection Method (see Figure~\ref{fig:RelatedCRLZSL}). The Projection
method provides insights for labeled instances from an unseen
class by projecting both the instance’s feature space and the
semantic space prototype to a shared space \cite{wang2019survey}.

Table~\ref{tab:relatedWorkTable}  shows existing Zero-Shot Learning (ZSL) models that were compared with our model in the evaluation section.
Most of the recent work includes two kinds of semantic spaces, namely Label-Embedding Spaces  \cite{wang2019survey,xu2019complementary} and Attribute Spaces \cite{wang2019survey} (also known as Probability Prediction Strategy \cite{xu2019complementary}).  Label-Embedding Spaces focuses on learning a projection strategy that connects image semantic features to labels, in which labels are represented a high dimensional embedding using Word2Vec \cite{mikolov2013distributed} or Glove \cite{pennington2014glove}. Image Features in Label-Embedding Spaces are typically learned from Convolutional
Neural Networks \cite{akata2016multi,akata2015evaluation,frome2013devise,Fu_2015_CVPR,romera2015embarrassingly,xian2016latent}.
Attribution Spaces or Probability Prediction initially focuses on pre-training attribute classifiers based on source data \cite{xu2019complementary}, 
where an attribute is defined as a list of terms describing various properties of given a class \cite{wang2019survey}. 
Each attribute forms the dimensions of class; value is typically given if a class contains the attribute or not \cite{lampert2013attribute,norouzi2013zero,yu2019meta}.

Pure Image Representation Space-based ZSL is rarely observed, one of the very first works was to use Image Deep Representation and Fisher Vector for the ZSL Project method
\cite{wang2017alternative}  and an extension of this approach was used to create unsupervised domain adaptation \cite{wang2019unifying}. Zhu et al. used the
partial image representation method to achieve a universal representation for action recognition \cite{zhu2018towards}.

\begin{table*}[]
    \centering
     \caption{Related Work: Zero-Shot Learning Methods}
    \renewcommand{\arraystretch}{1.2}
    \begin{tabular}{c|c|c}\hline
         {\bf Approach} & {\bf Semantic Space} & {\bf Zero-Shot Learning Method}  \\ \hline
       CRL Model (ours) & LSS - Image Representation Space & Instance-Based (Projection Method) \\ 
        Deep-WMM-Voc \cite{fu2019vocabulary} & LSS - Label Embedding Space & Classifier-Based\\
        SAE \cite{kodirov2017semantic} & ESS- Attribute Space \& LSS-Label Embedding Space & Classifier-Based\\
        ESZSL \cite{romera2015embarrassingly} & ESS- Attribute Space & Classifier-Based\\
        Deep-SVR \cite{lampert2013attribute}\cite{farhadi2009describing} & ESS - Attribute Space & Classifier-Based \\
        Embed \cite{zhang2017learning} & LSS - Label Embedding Space& Instance-Based \\
        ConSE \cite{norouzi2013zero} & LSS - Label Embedding Space & Instance-Based \\
        DeViSE \cite{frome2013devise} & LSS - Label Embedding Space & Classifier-Based\\
        AMP \cite{Fu_2015_CVPR} & LSS - Label Embedding Space & Classifier-Based\\\hline
    \end{tabular}
    \label{tab:relatedWorkTable}
\end{table*}

\begin{figure*}[h!]
\centering
\begin{tabular}{cc}
\multicolumn{2}{c}{\includegraphics[width=0.8\linewidth]{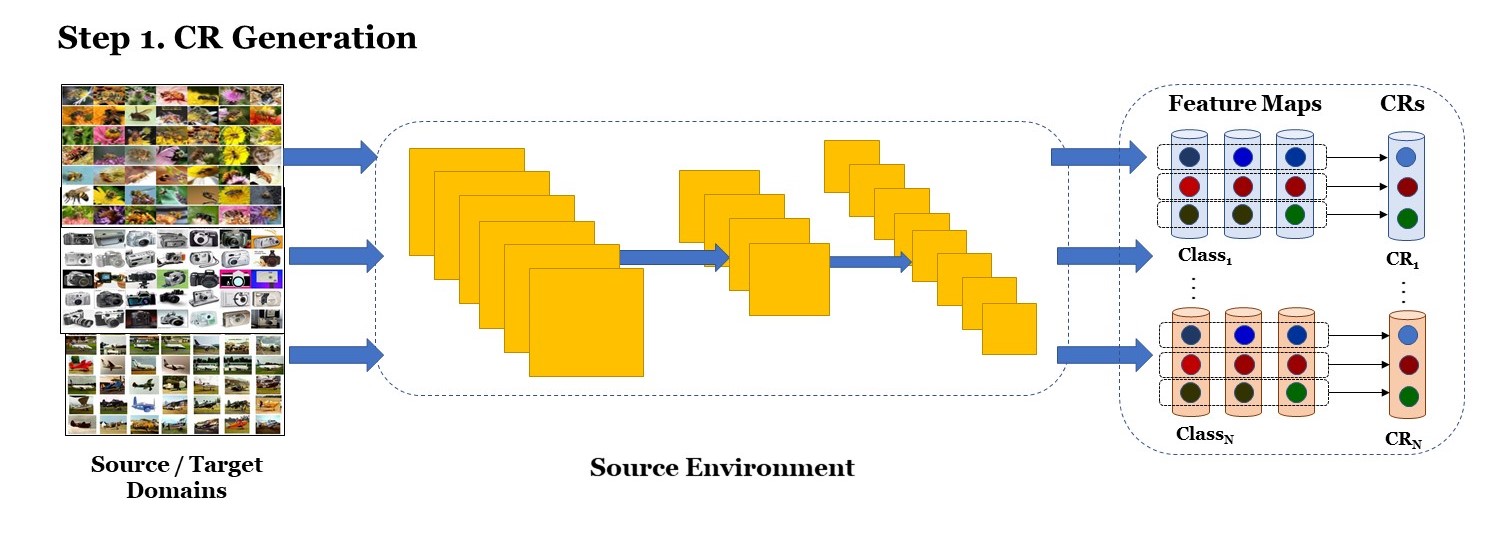}} \\
\multicolumn{2}{c}{\includegraphics[width=0.8\linewidth]{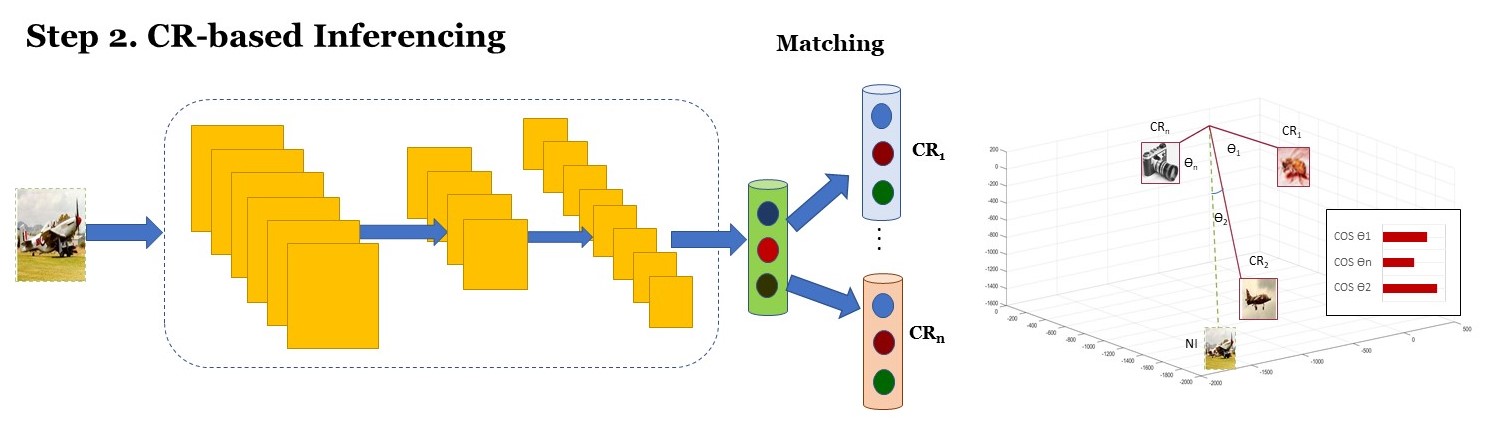}}
\end{tabular}
\caption{Class Representative Learning (CRL) Architecture}
\label{fig:Architecture} 

\end{figure*}

\section{Class Representative Learning Model}

The significance of the Class Representative Learning (CRL) model is its competence to project the input data to a global space that is specified 
by the activation of neurons in the pre-trained model such as CNN. The space of the CRL model is similar to the universal representation proposed 
by Tamaazousti et al. \cite{tamaazousti2019learning}, where visual elements in the configuration (e.g., scale, context) can be encoded universally for transfer learning. The fundamental concept of the CRL model is its ability to create the representatives of class in parallel and
independently without depending on other classes. 

As shown in Figure \ref{fig:Architecture}, the CRL model is composed of two primary components such as CR Generation and CR-based Inferencing. 
The model used to evaluate the CRL model is the Inception-V3 model that was pre-trained with ImageNet-1K \cite{ILSVRC15}. The pre-trained model is the  {\it Source} environment for the CRL model where no learning is happening, but the  {\it Source} environment was mainly used as a reference standard for producing a feature vector of the input data in space. Figure~\ref{fig:Architecture}(a) shows the {\it Source}  environment (i.e., Pre-trained model), and Figure~\ref{fig:Architecture}(b) shows the inferencing process with CRs on how a new image is projected on the  {\it Source}  environment and is mapped it on to the CRs for classification.

\subsection{Problem Setup}
Assume the given source data $D_s = \{ x_n,y_n\}^{m_s}_{n=1} $ of $m$ labeled points with a label from the source class ${1,\dotsc,S}$, where $x_n \in \mathbb{R}$ is the feature of the $n^{th}$ image in the source data and $y_n \in \mathbb{S}$ where $\mathbb{S}=\{1,\dotsc,C_s\}$ is the set of source classes. 
The target data is represented as $D_t = \{ x_n, y_n \}^{m_t}_{n=1}$ classes where $y_n \in \mathbb{T}$. The target classes $\mathbb{T}$ are represented as $\{C_s+1,C_s+2,\dotsc,C_t\}$ where $C$ total set of classes is $C = \mathbb{S} \cup \mathbb{T}$, where the total number of classes is $C_s+C_t$. For each class $c \in \mathbb{S} \cup \mathbb{T} $, has a Class Representative $CR(c)$ which is the semantic representative of class $c$. 
Furthermore, the source label $\mathbb{S}$ and the target label $\mathbb{T}$ are considered such that $\mathbb{S} \cap \mathbb{T} = \emptyset$.  For simplicity, the source and target datasets  have overlapped  labels but these overlapped classes are considered distinct.
In the CRL model, the source data are considered as {\it seen}, and the target data are {\it unseen}. In other words, the target data are not used in the learning process. Table~\ref{tab:notation} summarizes all the symbols and notations used in the CRL model.

The goal of the CRL model is that given a new test data $x^*$, the model will classify it into one of the classes $y^*$ where $y^* \in C$. 
The CRL model defines a universal problem for a classification approach as well as a traditional ZSL approach as follows:
\begin{itemize}
    \item Classification (CL) Approach: $y^* \in \mathbb{S}$
    \item Zero-Shot Learning (ZSL) Approach: $y^* \in \mathbb{S} \cup \mathbb{T}$
%    \item Generalized Zero-shot Learning $y^* \in \{\mathbb{S} , \mathbb{T} \}$
\end{itemize}

In both models, there are no dependencies among CRs.
The difference between these two models is in the properties of the inference. 
If the CR of the target set was introduced, then it would be {\it CL}, 
and if both CRs of the source set and the target set are introduced, then it would be {\it ZSL}.   

\begin{table}
    \renewcommand{\arraystretch}{1.2}
    \caption{Formal Symbol and Notations in the CRL model}
    \centering
    \begin{tabular}{|c|c|}
        \hline
        Notation  & Description \\
        \hline
        $D_s \& D_t$ & Source and Target Domain \\
        $m_s \& m_t$ & \#Data Points from Source and Target, respectively \\
        $C_s \& C_t$ & \#Classes from Source and Target, respectively \\
        $\mathbb{S} \& \mathbb{T}$ & Source and Target Label Set \\
        $C$ & \#Classes \\
        ${x}$ & Feature Vector of Labeled Data Point \\
        ${y}$ & Label of Data Point \\
        ${j}$ & \#Neurons in a Given Activation Layer \\
        $AFM_s$   & Activation Feature Map  $\{ b_1, b_2, \ldots, b_j \}$\\
        $CR(c)$ &  Class Representative of Class $c$ where $c \in C$\\
        $CR(c)$ & Class Representative Set $\{CR_c^1,CR_c^2,\dotsc,CR_c^n\}$\\
        ${x^*}$ & Unlabeled Data Point \\
        $CRC$ & Class Representative Classifier \\
        $CRFS^n$ & Class Representative Feature Space \\
        $n$ & Dimensions of the $CRFS$ \\
        \hline
    \end{tabular}
    \label{tab:notation}
\end{table}

\subsection{CR Definition and Property} \label{Section_CR_Definition}

\noindent {\bf Definition 1:} {\it Activation Feature Map}\\ 
Activation Feature Map (AFM) is a vector of features extracted from a base model that will be defined by a pre-trained model for any given instance.
For a given input, AFM  represents the features  that are defined by the activation of neurons in the base model. 
The AFM dimensionality is the number of neurons in the selected layer of the base model. In other words, it is the number of distinct neuron activating neurons occurring in the corpus. 
The $n$ dimensional AFM  forms the basis for the Semantic Space that is defined in Zero-Shot Learning (ZSL). 
%In CR, vector operations was used to classify a new input by comparing the CRs with queries.

\noindent{\bf Definition 2:} {\it Class Representative} \\ 
Class Representative (CR) is a representative of $K$ instances in a single class.
The Activation Feature Map of the CR is a unique characteristic pattern of visual expression that occurs as a result of the deep learning process 
in Convolutional Neural Networks (CNN). Thus, CR is an abstraction of instances of a class by computing an aggregation of the average mean vectors 
of the AFM for the $K$ instances. The CR characterizes a class and differentiates  one class against another. 
%One class may have multi-modality signatures (image, voice, behavior, text, etc). 

The Class Representatives $CR(c)$ for Class $c$ is represented as $\{CR_c^1,CR_c^2,\dotsc,CR_c^n\}$ with $n$ dimensions.
Each dimension corresponds to a separate feature. If a feature occurs in CR, its value in the vector is non-zero. 
%Several different ways of computing feature weights have been developed. 

\noindent{\bf{Definition 3:}}{ \it Class Representative Classifier} \\
A Class Representative Classifier $CRC: I^d \rightarrow C$ maps an input image space $I^d$ of the dimension $d$  
to the Class Representative Feature Space $CRFS^n$ of the dimension $n$ to classify it to Class $C$.  
(CRFS is defined in Definition 4).
$CRC$ is defined as a composition of two functions as shown in Equation~\ref{eq:CRC}.

\begin{equation} \label{eq:CRC}
\centering
\begin{split}
    CRC = L(S(.)) \\
    S : I^d \rightarrow CRFS^n \\
    L : CRFS^n \rightarrow C
\end{split}
\end{equation}

The CRC model first maps the Input Space $I^d$ to Class Representative Feature Space $CRFS^n$ with $n$ dimensions. 
$CRFS^n$ further maps into Class $C$. The mapping function $S$ represents the source environment, 
which aids the transformation of the input data into the Feature Space. (The source environment is further discussed in Section~\ref{AFM}).
For example, the input of a  {\it dog} image with dimensions [299x299x3] 
is mapped into Class Representative Space (as defined in Definition 4). Then, 
the CRFS will be labelled with the class {\it dog} through the classification process 
\cite{palatucci2009zero,fu2019vocabulary,fu2016semi}.

\noindent{\bf{Definition 4:}} {\it Class Representative Feature Space} \\ \label{AFM}
Class Representative Feature Space (CRFS) is a  $n$ dimensional semantic feature map 
in which each of the $n$ dimensions represents the value of a semantic property. These properties may be categorical
and contain real-valued data or models from deep learning methods \cite{palatucci2009zero}.
The Class Representative Feature Space  represents  $n$ dimensional representative features as a form of the Activation Feature Map (AFM).

%{\bf \Large the following section "Source Environment" should be defined here? what about introducing this before the CR definition and property section? \it{Function $S$ in the previous section}}

The design of the CRFS is based on the equations defines in \cite{raina2007self}.
The data points from $D_t$ $\{(x^{(1)},y^{(1)}, \ldots, (x^{(m_t)},y^{(m_t)}\}$ with each $x^{(i)} \in \mathbb{R} $ and $y^{(i)} \in \mathbb{T}$ (as shown in Equation~\ref{ActivationLayer2}.
A set of the mean of the base features is defined as $x^{(i)}$ 
%The labeled target data points can be defined as $D_t = \{(x^{(1)},y^{(1)}), (x^{(2)},y^{(2)}), ... , (x^{(m_t)},y^{(m_t)})\}$ as in Equation~\ref{ActivationLayer2} where $m_t$ is the number of data points. 
Note that the data points can also be defined in $D_s$, that will be used in CRFS to understand both source and target domains (refer to Section~\ref{SourceandTargetMap}).

\begin{equation}
%\begin{aligned}
\centering
\begin{split}
    D_t = \{ x^{(i)}, y^{(i)} \}_{i=1}^{m_t} \\
    \hat{a}(x^{(i)}) = arg min_{a(i)} || x^{(i)} - \sum_{j} a_j^{(i)} b_j ||_2^2 + \beta || a^{(i)}||_1\\
\end{split}
%\end{aligned}
 \label{ActivationLayer}
\end{equation}
\begin{equation}
\hat{D_t} = \{ (\hat{a}(x^{(i)}), y^{(i)} \}_{i=1}^{m_t}
\label{ActivationLayer2}
\end{equation}

CRFS is created based on the base vectors $b = \{b_1, b_2, \ldots , b_s\}$ with each $b_j \in 
R_n$. The base vector $b$ is generated in the source environment using $D_s$. The activation $a = \{a^{(1)}, . . . , a^{(k)}\}$ with each $a^{(i)} \in R_s$ forms the semantic property of CRFS. Each dimension of an activation vector $a^{(i)}_j$ is the transformation of input $x^{(i)}_u$ using the base $b_j$. The number of bases $s$ can be much larger than the input dimension $n$. The transformed target data points $\hat{D_t}$ will be input for the Class Representative Generation. 

Since each pair of $(x^{(i)}, y^{(i)})$ is independent of each other, our algorithm was designed with the CRCW  (Concurrent Read Concurrent Write)  model 
which allows parallel computing, including I/O, with the shared memory and processors.
The CRL's operation time is proportional to the number of the selected CRs across all processors.
The CR Generation will be proportional to the input set per CR independent of the number of classes in a given dataset.
The CR-based inferencing will be proportional to the number of CRs in a given model.

\subsection{CR Generation} \label{Section_CR_Generation}

Class Representatives (CR) are generated using the nearest prototype strategy by aggregating feature vectors.
The nearest mean feature vector with instances of the given class, i.e., Class Representatives, is computed for each class. 
Specifically, the average feature mean operation was used to summarize the instances of classes. 
For each class, the instances of each feature in the feature maps (e.g., 12K for a CNN layer) are aggregated 
into a simple mean feature in order to create its CR. The CR is an aggregated vector of the mean features for all the features in the feature maps.

For the Class Representative Generation, we considered the transformed Target Dataset $\hat{D_t}$ as the input (as shown in Equation~\ref{ActivationLayer}). As we emphasis on the parallelism and independence, we considered the individual activation vector $\hat{a}(x^{(i)})$ such that $y^{i} = c$, that will be used in formulating the CR as shown in Equation~\ref{CR Generation}.
\begin{equation}
\label{CR Generation}
\begin{split}
    CR(c) = \{CR^1, CR^2, ... , CR^n\} \\
    CR^j = \dfrac{1}{N_c} \sum_{j=1}^{N_c} \hat{a}(x^{(i)}) \\
\end{split}
\end{equation}

where $j$ ranges from $1$ to $n$ representing the feature dimensions, $c$ is the class of the input image, and $N_c$ is the number of data points for the class $c$. Class Representative of the given class $c$ is represented as the group of CR features values $CR^j$ where $j$ ranges from 1 to $n$ feature dimensions. $CR^j$ is generated from the mean of AFM (refer to Equation \ref{ActivationLayer}) of every input image in a given class  $c$ as shown in Equation \ref{CR Generation}.  The CR Generation can be conducted in parallel so that each CR of class independently can be generated. The parallel  algorithm with CRCW (as explained in Section \ref{AFM}) was implemented with Spark's broadcast variables for the CR Generation. 

\subsection{CR Feature Space: Source and Target Mapping} \label{SourceandTargetMap}
The Class Representative (CR) mapping is a variation of Multi-Discriminative Problem network \cite{tamaazousti2019learning}. 
This method is attempting to universalize a method that combines different but complementary features learned on different problems.
 The source domain problem is defined as the $DP^s$ in the class when the Convolution Neural Network 
 assigns to the input image the label corresponding to the classes provided by the source domain $D$.  %(here ImageNet1k) %. 
Similar to what is described in Pan et al. \cite{pan2010survey}, we define our source and target domain based two aspects, namely Class Representative Feature Space (CRFS) and Marginal Distribution.

% target domain problem $DP^t$ such that the domain has $t$ or more new image classes than what $DP^s$ is trained on. 

\begin{equation}
\label{DomainDefintion}
\begin{split}
    D_s = {CRFS_s, P(CR_s)} \\
    D_t = {CRFS_t, P(CR_t)}  \\
\end{split}
\end{equation}

As shown in Equation~\ref{DomainDefintion}, the CR source domain $D_s$ is a two-element tuple consisting of Source CR Feature Space $CRFS_s$ and Probability Distribution Function of CR $P(CR_s)$,
where $CR_s$ is Class Representatives from the source domain. The CR target domain $D_t$ is also defined as a two-element tuple consisting 
of Target CR Feature Space $CRFS_t$ and Probability Distribution Function of CR, $P(CR_t)$, where $CR_t$ Class Representatives were generated from the target domain. 

% {\bf\Large HERE: show where $D_s$ and $D_t$ can be used.}

\begin{equation}
\label{Eq:kolmogrov}
\begin{split}
    %Kolmogrov-Smirov statistic 
    D^* = \max_x\abs{P(CR_s) - P(CR_t)} \\
    P(CR_s)   \forall    CR_s \in CRFS_s \\
    P(CR_t)    \forall    CR_t \in CRFS_t \\
    \end{split}
\end{equation}

As shown in Equation~\ref{Eq:kolmogrov}, $P(CR_s)$ and $P(CR_t)$ are the distribution functions based on the probability distribution for the source and target domain CRs respectively. $D^*$ shows the Kolmogorov-Smirnov distance  between the source CR distribution and target CR distribution, i.e., it is computed as the max of distance between $P(CR_s)$ and $P(CR_t)$. 
    
We use Two-Sample Kolmogorov-Smirnov Test (KS-Test) as a simple test for measuring the differences of the  distributions of two sets, such as a sample and a reference probability distributions. Equation~\ref{Eq:kolmogrov} computes the distance $D^{*}$ between the medians of the Source Distribution $P(CR_s)$ and Target Distribution $P(CR_t)$. The distance $D^{*}$ is the indicator that would be used to measure the CR similarity between $P(CR_s)$ and $P(CR_t)$.  A larger distance $D^{*}$ yields less accurate transfer learning for the target domain. 

\subsection{CR-based Inferencing} \label{Section_CR_Inferencing}

The CR-based inferencing is a mapping between the input and Class Representatives (CRs) and label it with a class.
The CR-based inferencing can be done in parallel since the CRs are independent of each other.

\begin{equation}
\centering
    \label{CosWithImageWithCR}
    \begin{aligned}
     \mathrm {cos} (CR(c), NI)& ={\frac  {CR(c) \cdot NI }{\left\|CR(c) \right\|\left\|NI \right\|}}\\
    & = {\frac {\sum _{i=1}^{N}x_{i,c}x_{i,ni}}
     {{\sqrt {\sum _{i=1}^{N}x_{i,c}^{2}}}
     {\sqrt {\sum _{i=1}^{N}x_{i,ni}^{2}}}}}
    \end{aligned}
  \end{equation}
Here are the steps for the CR-based inferencing.
Given a new input is vectorized in the source environment to the Class Representative Feature Space (CRFS), $NI = \hat{a}(x^{(i)})$, as shown in Equation~\ref{ActivationLayer}. The cosine similarity between the new input ($NI$) and Class Representatives for class $c$ ($CR(c)$), where $c \in C$ can be computed using Equation~\ref{CosWithImageWithCR}.  The CRL Model assigns the new input with the label associated with Class $c$ that has the highest cosine similarity score. The higher cosine similarity score indicates the closeness between the Class Representative $CR(c)$ and the new input ($NI$) in the Class Representative Feature Space (CRFS).

  \begin{equation}
  \label{ArgMaxCosine}
        \hat{c} = \argmax_{c \in C}  \{\mathrm {cos} (CR(c), NI)\}
  \end{equation}

As shown in Equation~\ref{ArgMaxCosine}, the label for the new input from CRL Model $\hat{c}$ is predicted by selecting the class from all classes $C$ that has the highest cosine similarity to the new input. The CRL model will conduct inferencing  
by matching the new input against the available CRs and label it with a class having the highest cosine similarity score.

\section{Implementation and Experimental Design}

The experiments on the Class Representative Learning (CRL) model have been conducted on ImageNet-1K as a source domain and CIFAR-100, CalTech-101, and CalTech-256 as a target domain. The {\it source} environment, i.e. the pre-trained model from the source dataset (ImageNet-1K), with three different deep learning networks, such as Inception-V3 \cite{szegedy2016rethinking}, ResNet-101 \cite{he2016deep}, and VGG-19 \cite{simonyan2014very}.

\subsection{Implementation}\label{ED-Implementation}

\subsubsection{System and Library Specifications}
The Feature extraction was implemented on a single GPU, which is Nvidia GeForce GTX 1080 (with 12GB GDDR5X RAM) on MATLAB 2018b version. The CR Generation and CR-based Inferencing were implemented using Spark 2.4.3 version \cite{ApacheSp82:online}. The parallel and batch process was conducted through RDD based parallelism on a single CPU with 4GHz Intel Core i7-6700K (quad-core, 8MB cache, up to 4.2GHz with Turbo Boost) and 32GB DDR4 RAM (2,133MHz)  (i.e., local parallelism of 4 cores). 

\subsubsection{Models Specification}
As described in Section~\ref{AFM}, the {\it source} environment provides the feature space for the CRL model. The Inception-V3 model was predominantly used as the source environment (pre-trained with ImageNet-1K) for the CRL experiments. The Inception-V3 model was obtained from MATLAB's Pre-trained Deep Neural Networks \cite{Pretrain97:online}. The layer information of the Inception-V3 model is shown in Figure~\ref{fig:Activations_LayerWise}. The feature extraction has been conducted class by class as a form of parallel processing to build a CR for each class. For some experiments,  ResNet-101 and VGG-19 extracted from MATLAB were also used as our source environments. The last convolution layer from the source environments was considered for Feature Extraction. 

The CR Generation was implemented in parallel with Spark's Resilient Distributed Datasets (RDDs), which is a collection of features partitioned across the nodes of the cluster. The batch in this context was defined while keeping CR independence of each class for the CR Generation and CR-based Inference.

\subsection{Datasets}\label{ED-Datasets}

We have conducted the experiments with the CRL model using four datasets according to the three transfer learning types defined in Day et al.  \cite{day2017survey}, 
i.e.,  Homogeneous Transfer Learning (HOTL), Heterogeneous Transfer Learning (HETL), and Negative Transfer Learning (NTL). 
\begin{itemize}
    \item Homogeneous transfer learning happens when the source and target feature spaces have the same attributes, labels, and dimensions
    \item Heterogeneous transfer learning happens when the source and target domains may share no features or labels and dimensions of the feature space may differ as well
    \item Negative transfer learning happens when the target domain's performance negatively impacts due to knowledge transfer from the source domain. The negative transfer learning is generally found when the source domain has very little common to the target domain. 
\end{itemize}

\begin{figure}[h]
    \centering
    \includegraphics[width=8cm]{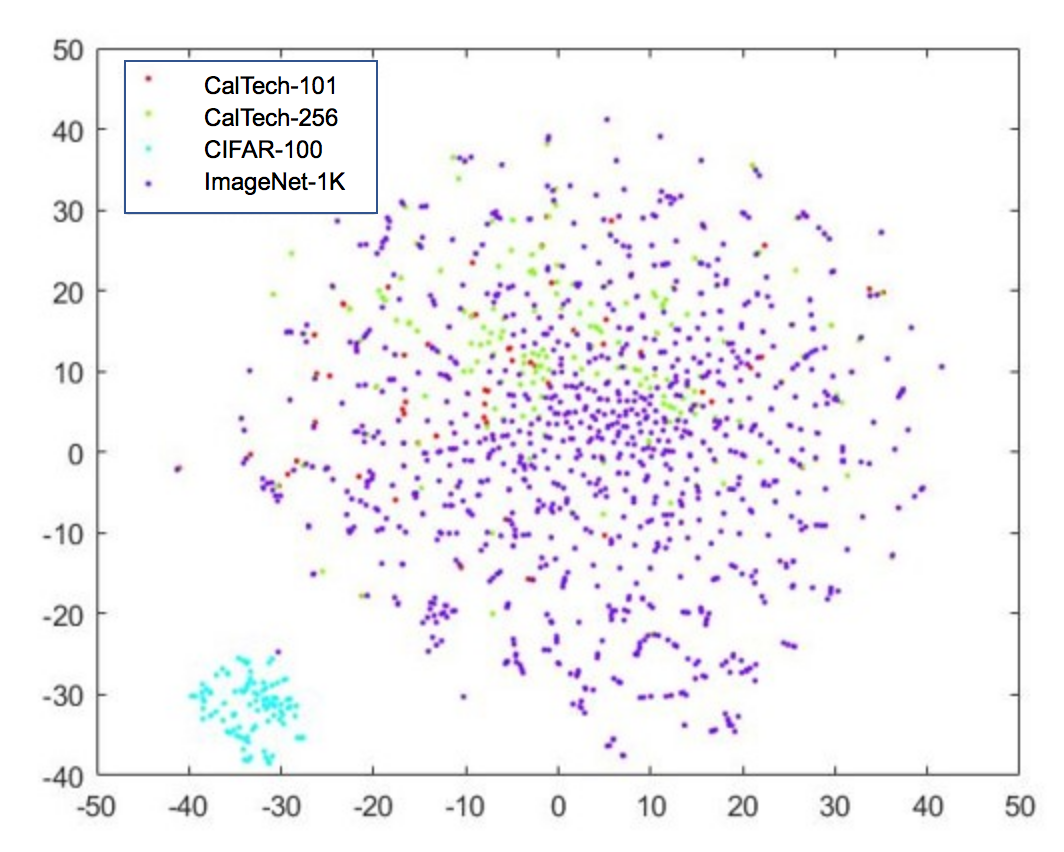}
    \caption{t-SNE Visualization of Class Representatives}
    \label{fig:TSNE}
\end{figure}

\begin{figure}[h]
    \centering
        \begin{tabular}{cc}
           \includegraphics[width=0.45\linewidth]{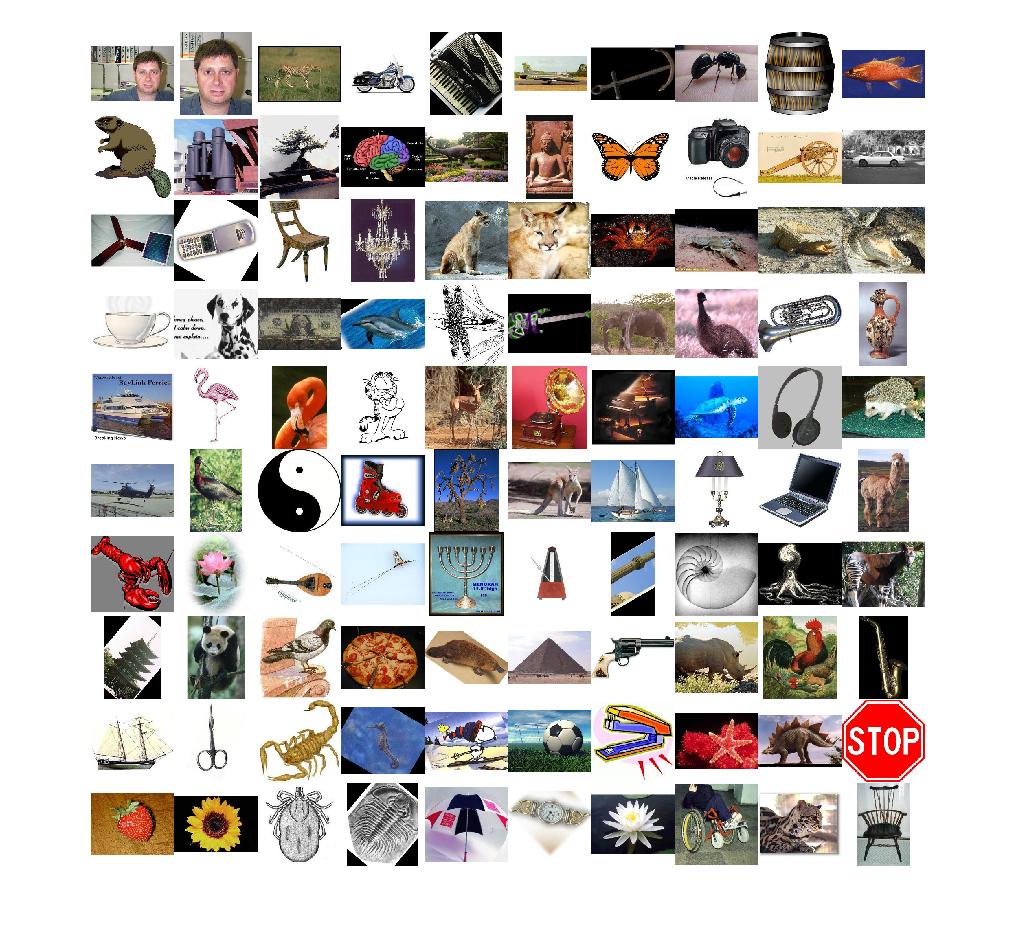}  &  \includegraphics[width=0.45\linewidth]{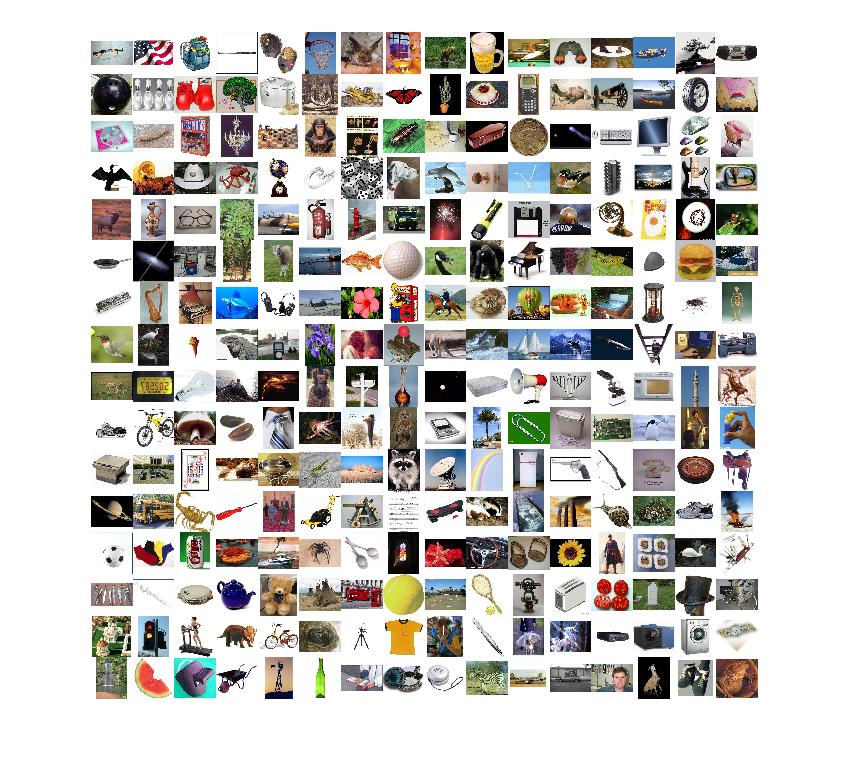} \\
           CalTech-101 & CalTech-256 \\
           \includegraphics[width=0.45\linewidth]{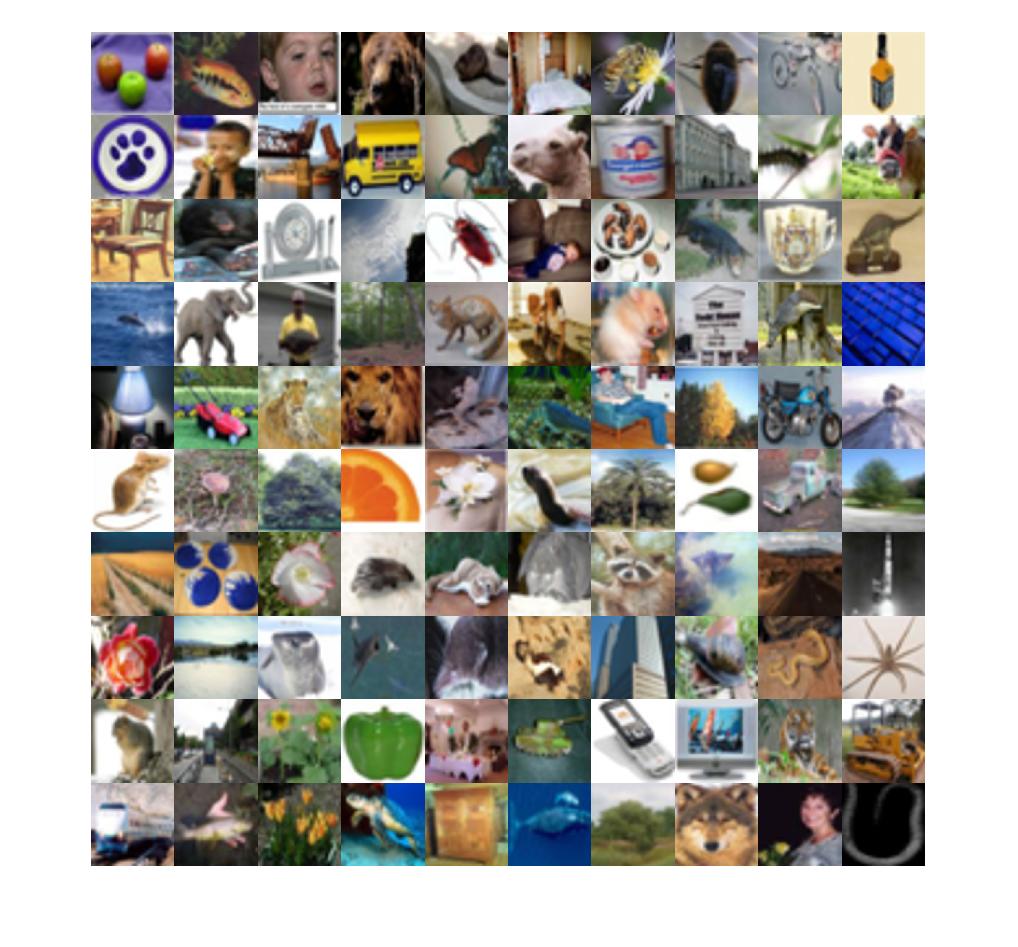} &
           \includegraphics[width=0.45\linewidth]{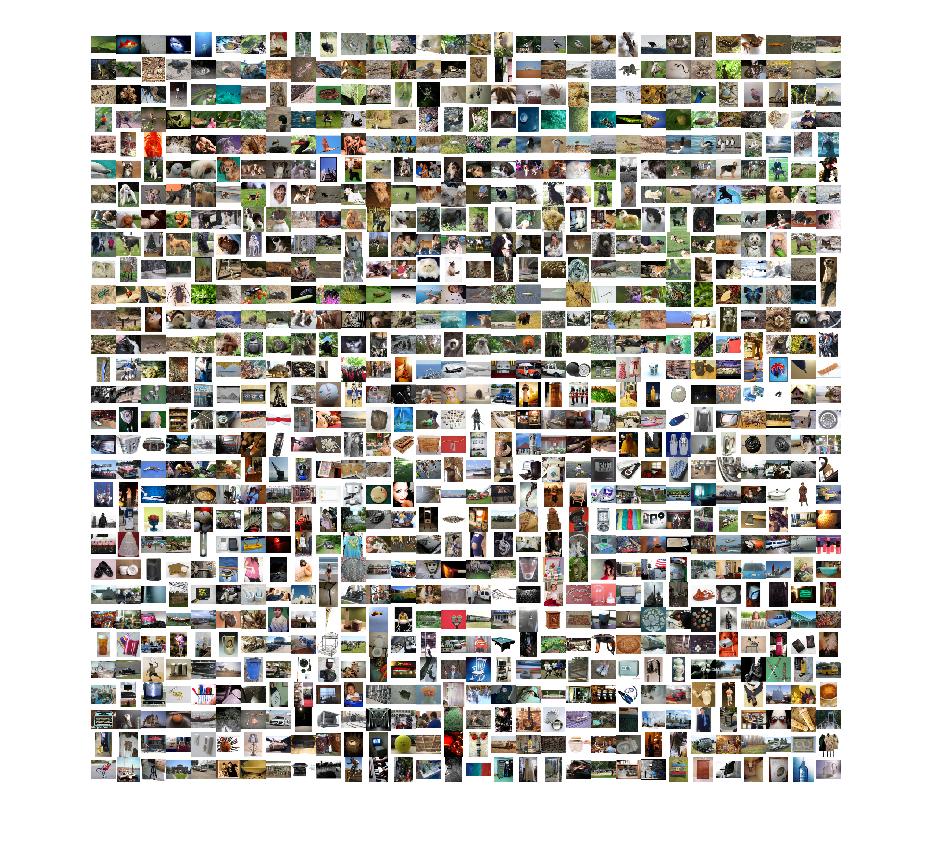} \\
           CIFAR-100 & ImageNet-1K\\
        \end{tabular}
        \caption{Benchmark Datasets: CalTech-101 \& CalTech-256 \& CIFAR-100 \& ImageNet-1K}
    \label{fig:ExampleOFDataset-2}
\end{figure}

The four datasets include ImageNet-1K, CalTech-101, CalTech-256, and CIFAR-100 (as shown in Table \ref{table:Dataset} and Figure \ref{fig:ExampleOFDataset-2}). 
Figure~\ref{fig:TSNE} shows the four datasets that were used for the CRL's transfer learning.

The source environment is built on the ImageNet-1K dataset that is the Homogeneous Transfer Learning (HOTL) type (the same label set and the same attributes).
The transfer learning with CalTech-101 and CalTech-256 are the Heterogeneous Transfer Learning (HETL) type, which was projected on the semantic space of the source domain with minimal distinction classes.  
The transfer learning with CIFAR-100 is the Negative Transfer Learning (NTL) type since the target domain data are projected on the semantic space that is quite distinct from the source domain. 
Although the CIFAR-100 is semantically relevant to other datasets, the CRL space of CIFAR-100 is divergent from the source space in terms of image modality, such as image quality and image size.  
The size of CIFAR-100 images is [32x32] while one of the source domain ImageNet-1K [400x400].
More specifically, the dimension of the source environment of Inception-V3 is [299x299]. 
In the experiment section, the details on the performance of these different transfer learning types will be discussed.

\begin{table}[h]
\renewcommand{\arraystretch}{1.2}
\caption{Benchmark Datasets}
\label{table:Dataset}
\begin{tabular}{|c|c|c|c|c|}
\hline
\textbf{Domain} & \textbf{Dataset} & \textbf{\#Class} & \textbf{\#Image} & \textbf{Image Size} \\ \hline
\textbf{\begin{tabular}[c]{@{}c@{}}Source \end{tabular}} & ImageNet-1K & 1000 & 1,281,167 & 400x400 \\ \hline
\multirow{3}{*}{\textbf{\begin{tabular}[c]{@{}c@{}}Target \end{tabular}}} & CalTech-101 & 101 & 8,677 & 300x300 \\ \cline{2-5} 
 & CalTech-256 & 256 & 30,608 & 300x300 \\ \cline{2-5} 
 & CIFAR-100 & 100 & 59,917 & 32x32 \\ \hline
\end{tabular}
\end{table}

\subsection{Experiments for Transfer Learning Performance}\label{ED-Domain}

Transfer Learning Performance in terms of Accuracy and Time or Space Requirements: state-of-the-art Transfer Learning vs. CR-based Classification with different datasets (CalTech-101, CalTech-256, ImageNet-1K, ImageNet-20K, CIFAR-100).

\begin{table*}[]
\renewcommand{\arraystretch}{1.2}
    \centering
     \caption{Source Domain and Target Domain: Cosine Similarity Median, Accuracy, Kolmogorov-Smirnov Test Scores\\  \footnotesize{ *CS: Cosine Similarity. A high CS Distance ($S_m - T_m$) and high KS Test Score indicate Negative Transfer Learning. \\
     State-of-the-Accuracy is from Inception-V3 Model (Refer Table~\ref{ResNet&Inception})}}
    \begin{tabular}{|c|p{1.3cm}|p{1cm}||c|p{1.3cm}|p{1cm}|p{1.3cm}||p{1.5cm}|p{1.5cm}|p{1.5cm}|}
        \hline
        \multicolumn{3}{|c||}{\bf Source Domain}& \multicolumn{4}{|c||}{\bf Target Domain} & \multicolumn{3}{|c|}{\bf Source \& Target Comparison}\\\hline
         Dataset  & CS Median ($S_m$) & Accuracy ($S_a$) & Dataset  & CS Median ($T_m$) &  Accuracy ($T_a$) & State-of-the-art Accuracy ($SA_a$) & CS Distance ($S_m - T_m$) & KS Test Score  & Accuracy Change  ($SA_a - T_a$)\\ \hline
        ImageNet-1K& 0.434  & 73.9\%   & CalTech-101 &  0.398 & 93.9\% & 92.98\%  & 0.036 &0.1570 & +1\%\\ \hline
         ImageNet-1K & 0.434 & 73.9\% & CalTech-256 & 0.454 & 77.9\% & 77.6\% & 0.02 &0.0921 &  +0.3\% \\ \hline
          ImageNet-1K&  0.434 & 73.9\% & CIFAR-100& 0.734  & 57.9\%  & 76.2\%  & 0.3 &0.9125 & -18.3\%\\ \hline
          ImageNet-1K &  0.434 & 73.9\%  &ImageNet-1K & 0.434 & 73.9\% & 78.8\% & 0 & 0 & -4.9\% \\ \hline
    \end{tabular}
    \label{tab:ks-test}
%    \vspace{1mm}
 %   \footnotesize{ *CS- Cosine Similarity. Higher CS Distance ($S_m - T_m$) correlates to Negative Transfer Learning}
\end{table*}

\begin{table*}[h]
    \renewcommand{\arraystretch}{1.2}
    \centering
    \caption{Class Representative Distribution Statistics}
    \begin{tabular}{|c|c|p{1.35cm}|c|p{1.35cm}|c|p{1.35cm}|c|p{1.3cm}|}\hline
         & 	 \multicolumn{2}{|c|}{CIFAR-100} &  \multicolumn{2}{|c|}{CalTech-101} & \multicolumn{2}{|c|}{CalTech-256} & \multicolumn{2}{|c|}{ImageNet-1K}\\\hline
        &	Accuracy	& GCS & 	Accuracy & GCS & 	Accuracy & 	GCS	&  Accuracy & 	GCS \\\hline
        Mean Accuracy &	57.9\% &	0.74 &	88.8\% &	0.41 &76.0\%	& 0.46	 & 	73.8\% &	0.43 \\\hline
        %Median 	& 57.8\% &	0.74 &	91.7\% &	0.40 &		78.0\% &	0.46 & 75.4\% &	0.44 \\\hline
        Std dev	 &  15.2\% &	0.09 &	9.7\% &	0.13 & 15.6\% &	0.11	&	12.9\%&	0.10 \\\hline
        Range &		[17.7\%, 86.4\%] &	[0.44, 0.98]	& [55.6\%, 100\%] &	 [0.03, 0.92] &[29.1\%, 100\%]	& [0.08, 0.96] & [28.6\%, 96.5\%]	& [0, 0.99] \\\hline
    \end{tabular}
    \vspace{1mm}
    
     \footnotesize{GCS: Group Cosine Similarity}
    \label{tab:BenchmarkDataset}
\end{table*}

\begin{figure*}[h!]
\centering
\begin{tabular}{cc}
 
    \includegraphics[width=8.5cm]{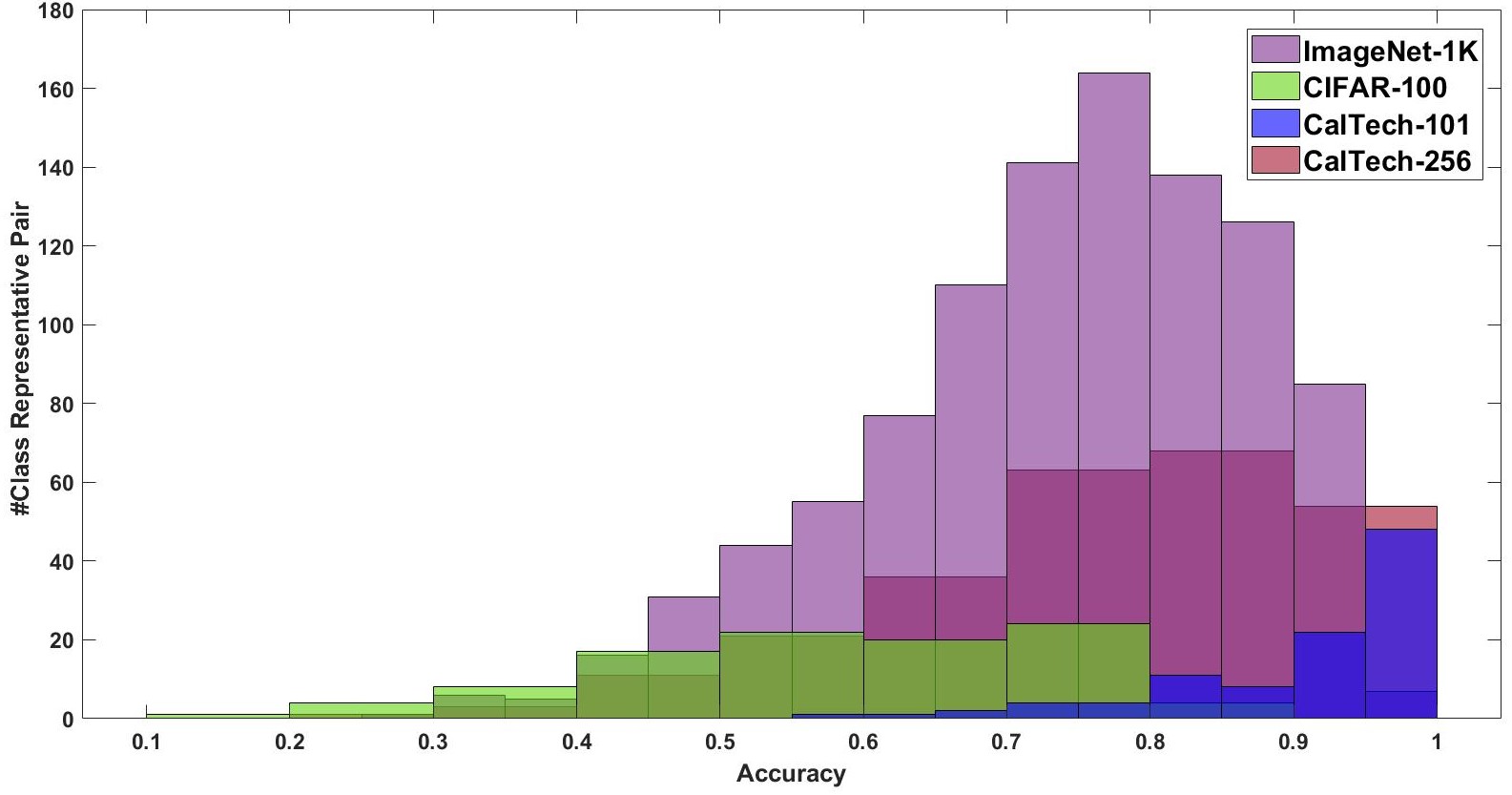}&
    \includegraphics[width=8.5cm]{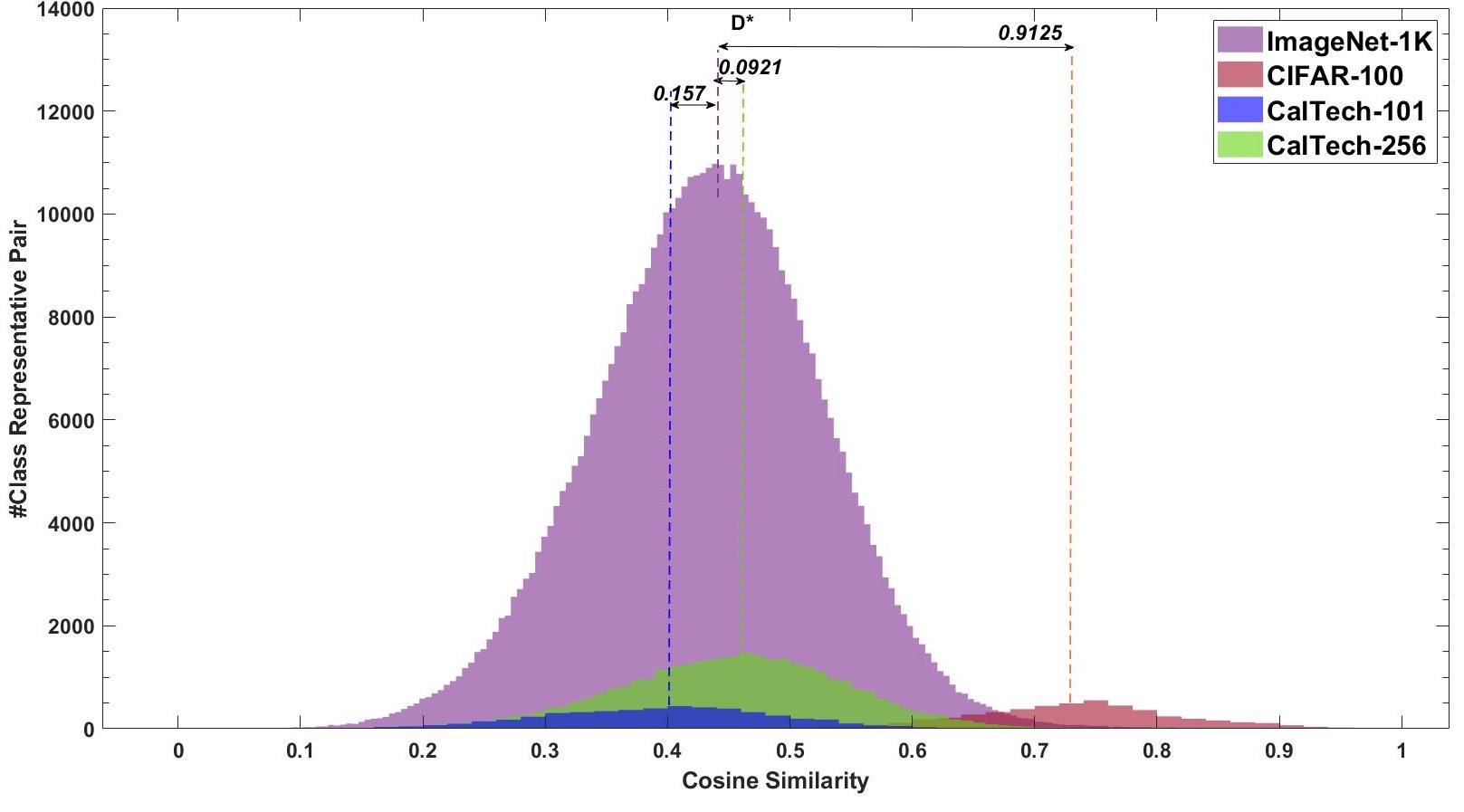}\\
%    \multicolumn{2}{c}{\includegraphics[width=15cm]{Images/SignatureImages/All_WithDist.jpg}}\\
\end{tabular}
\caption{Accuracy Distribution and Cosine Similarity Distribution}
\label{fig:Distribution} 
\end{figure*}

\begin{itemize}
% https://www.analyticsvidhya.com/blog/2017/06/transfer-learning-the-art-of-fine-tuning-a-pre-trained-model/
\item Case 1: Source Domain ($\mathbb{S}$): ImageNet-1K, Target Domain ($\mathbb{T}$): CalTech-101 or CalTech-256: The distance between the medians of both domains, SD and TD, is very small (i.e., $D^* \le 0.05$). In this case, the classification with the class representatives (CV), which are generated from the pre-trained model in SD with small data in TD, are as effective as the state-of-the-art models. 

\item Case 2: Source Domain ($\mathbb{S}$): ImageNet-1K, Target Domain ($\mathbb{T}$): CIFAR-100: The distance between the medians of both domains,$ \mathbb{S}$ and TD, is very big (i.e., $D^* \ge 0.3$) as well as the size of the data in TD is small.
\end{itemize}

Two-Sample Kolmogorov-Smirnov Test is used to determine whether the instances of any given class are distributed within a class. The class distribution would also be applied to determine if there are any data issues such as data labeling errors or noise data. Thus, we could estimate the class accuracy using the class distribution model even before the training.
Figure \ref{fig:Distribution} and Table \ref{tab:ks-test} shows the KS-Test results between the distribution of the source and target datasets.  If the KS-Test value is high, then the source model may not be suitable for the target domain. 

Figure \ref{fig:Distribution} demonstrates the similarity distribution in the feature space of the source and target datasets as well as their accuracy distributions. Accuracy distribution represents the histogram of class accuracy in a given dataset while using CR based classification. Cosine similarity distribution represents the cosine similarity between a CR Pair. Higher cosine similarity means the similarity between CRs is high. The cosine similarity distribution is using to compare the {\it source} dataset to {\it target} dataset. The comparison between CIFAR-100 and ImageNet-1K has the highest KS-Test value among the four different datasets.

The CRL model is used to understand the distribution of datasets and their performance. 
It also showcased the overall group cosine similarity where GSC of a class($c$) is defined as the sum of cosine similarity between $c$ and all other classes belonging to $C$ %(refer to Section~\ref{section_III_D} and Equation~\ref{Group_Cosine_Similarity}). 
Table~\ref{tab:BenchmarkDataset} shows the CR distribution statistics for the source and target domains in terms of their accuracy and group cosine similarity (GCS).
The results based on the CR-Inception-V3 as seen from Table~\ref{tab:BenchmarkDataset} are consistent with Figure~\ref{fig:TSNE}, 
t-SNE Visualization shows that CalTech-101 and CalTech-256 are overlapped with ImageNet-1K (source domain), while CIFAR-100 is in a long distance from the source domain.
The CalTech-101 shows the best mean accuracy and low cosine similarity. However, CIFAR-100 is limited by low mean accuracy and high cosine similarity. 
For the CIFAR-100 dataset, the accuracy of 57.9\%  is the least, and the group cosine similarity of 0.74 is the highest compared to the ones for all other datasets. 
The salient reason for the low accuracy of the CIFAR-100 dataset is mainly due to high cosine similarity and a huge distance from the source domain.
In summary, among the four datasets, CIFAR-100 performs the worst, and CalTech-101 performs the best.

\subsection{Classification Performance with Benchmark Datasets}\label{ED-Accuray}

In this paper, the  CRL model has been validated in terms of CR Feature Exaction and CR Generation as follows:
i) Feature extraction in terms of  CNN Network models (Inception-V3, ResNet-101) and CNN layers, 
ii) Feature representation and optimization such as (12K vs. 3K feature vector) and the number of training images (20,  30,  60,  100, and  All).
For most of the evaluation, CR-Inception-V3 version was considered. 

\subsubsection{Results for Architecture Selection in Feature Extraction}
The CRs are mainly dependent on the quality of the features extracted from the pre-trained CNN model.
The two popular pre-trained models such as Inception-V3  \cite{szegedy2016rethinking} and ResNet-101  \cite{he2016deep}  were used as the CR source environments and their performance was compared  in Table~\ref{ResNet&Inception}. We also evaluated to select the most suitable layer in these pre-trained models. For both the pre-trained models, we compared CRL with the original accuracy, as shown in the state-of-the-art approaches (\cite{szegedy2016rethinking,he2016deep,zagoruyko2016wide,tamaazousti2017mucale,shah2016deep}).

The accuracy for the datasets reported here is for the Top-1 accuracy of the model. Comparing the CRL model to the original model, it was observed that CalTech-101 has an increase of 1.56\%   in both the models. There was a significant decrease in the accuracy of The CIFAR-100 for both the models. This is likely due to the greater distance in a semantic space 
between the source domain (ImageNet-1K) and the target domain (CIFAR-100), as discussed in Section\ref{ED-Datasets}.
In the end, for the overall comparison of the models, the accuracies of the Homogeneous Transfer Learning (HTL) in Inception-V3  are better than the ones in ResNet-101. 
This comparison leads to the use of Inception-V3 as the source environment of the CRL model. 

\begin{table*}[h]
\renewcommand{\arraystretch}{1.2}
\centering
\caption{CRL Models based on Deep Learning Networks (Inception-V3 vs. ResNet-101)}
\label{ResNet&Inception}
\begin{tabular}{|c|c|c|c|c|}
\hline
Base Model & \multicolumn{2}{c|}{\bf Inception-V3} & \multicolumn{2}{c|}{\bf ResNet-101} \\ \hline
Comparison & CRL Model & State-of-the-Art Model \cite{szegedy2016rethinking} & CRL Model & State-of-the-Art Model \cite{he2016deep}\\ \hline
Parameter & 0 & 21.8M & 0  & 22.44M \cite{zagoruyko2016wide} \\ \hline
Feature & 3072 %12288 
& 2048 & 100352  & 2048\\ \hline
Shape & AvgPool({[}8x8x192{]})  & linear {[}1x1x2048{]} & {[}7x7x2048{]} & linear {[}1x1x2048{]} \\\hline
Layer & conv2d\_94 & 48 layers deep (347 layers) & res5c\_branch2c & 101 layers deep (316 layers) \\ \hline
ImageNet-1K & 74.07\% %73.93\% 
& 78.8\% & 68.4\% & 77.56\% \cite{zagoruyko2016wide}\\ \hline
CalTech-101 & 93.69\% %93.96\% 
& 92.88\% & 93.46\%  & 91.4\% \cite{tamaazousti2017mucale}\\ \hline
CalTech-256 & 77.78\% %77.87\% 
& 77.6\% & 76.76\% & 80.1\% \cite{tamaazousti2017mucale}\\ \hline
CIFAR-100 & 57.63\% %57.96\% 
& 76.2\%& 63.02\% & 72.77\% \cite{shah2016deep}\\ \hline
\end{tabular}
\end{table*}

\subsubsection{Results for Layer Selection in Feature Extraction}

\begin{figure*}[h!]
    \centering
    \includegraphics[width=1\linewidth]{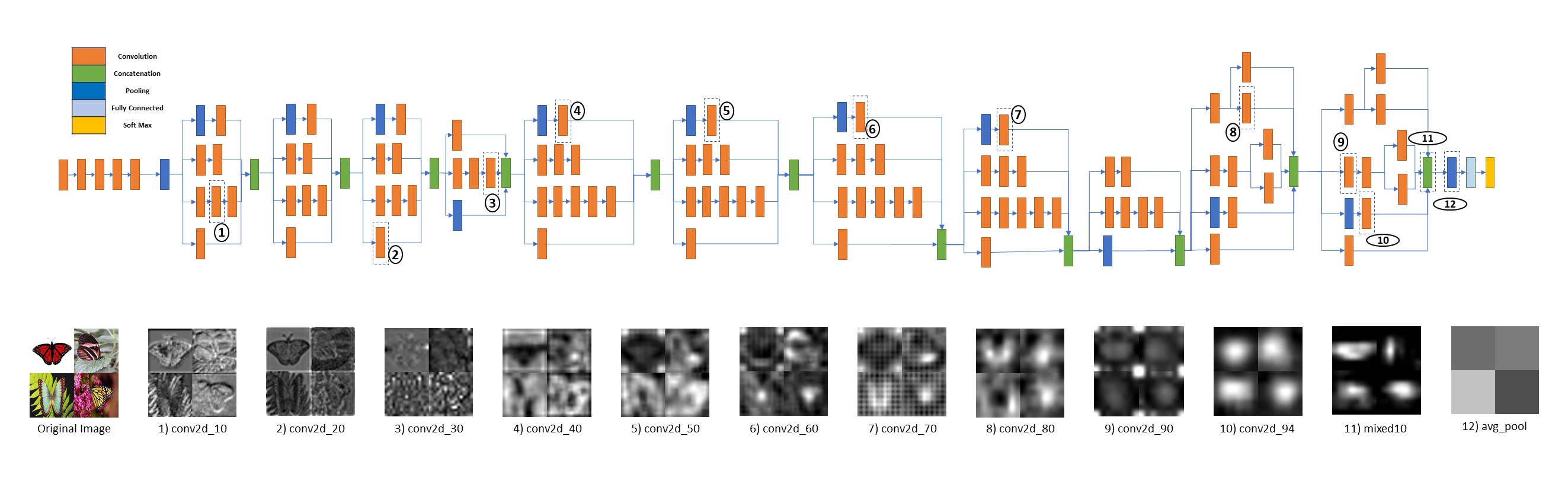}
    \caption{Inception-V3 Activation Visualization Layerwise}
    \label{fig:Activations_LayerWise}
\end{figure*}

From Inception-V3, the most suitable layer for the Feature Extraction was identified by the layer-wise experiment. 
As shown in Table~\ref{table:layerwise}, the accuracy evaluation was conducted with the models built using the features from the selected layer. 
For this comparison, the twelve layers (including ten different convolution layers, final concatenated convolution layer, and final average pooling layer) were considered. These layers are indexed, as shown in  Figure~\ref{fig:Activations_LayerWise}.

The best layer was determined in terms of the feature size and the accuracy of the model. Table~\ref{table:layerwise} shows the feature size, flatten feature size, and accuracy. For this evaluation, comparable datasets are considered to evaluate the effectiveness of the CR-based classification. The Homogeneous Transfer Learning (HTL) was used to conduct non-biased feature analysis and layer selection. Layer 10 (conv2d-94) shows the best accuracy in CalTech-101, while Layer 12 (AvgPool) shows the best accuracy in CalTech-256. For the flatten feature size, the feature set of Layer~10 shows the highest accuracy compare to other layers.

\subsubsection{Results from Feature Reduction}

Once the CR feature map is generated, the CRL model might be required to compress it for mobile deployment.
In this paper,  we have applied three different sampling techniques for the model compression: (1) Max Pooling, (2) Average Pooling, and (3) Min Pooling.
The Max Pooling is a sample-based discretization technique that is widely used in Deep Learning. 
The objective of the Max Pooling is to reduce the feature map's dimensionality by applying the max operation to features contained in the sub-regions of the feature map. The initial input of the CR feature map, such as $X*X$ matrix  (e.g.,  8*8), will produce to $Y*Y$ matrix (e.g., 4*4) using a Z*Z filter (e.g., 2*2).  A stride of S  (e.g., 2) controls how the filter operates around the input matrix by shifting S units at a time without any overlap regions. 

For each of the regions represented by the filter, the max of that region is computed to create the output feature map, in which each element is the max of a region in the original input. 
The Average Pooling and Min Pooling are very similar to the Max Pooling; the only difference is to utilize a different operation 
such as average and min operations for the feature map reduction.

\begin{table*}[h]
\renewcommand{\arraystretch}{1.2}
\centering
\caption{Inception-V3 Layerwise Accuracy on Similar Domain Datasets (SID)}
\begin{tabular}{|c|p{2cm}|p{2cm}|p{2cm}|p{2cm}|p{2cm}|p{2cm}|}
\hline
  % & Dataset      & \begin{tabular}[c]{@{}c@{}}Convolution \\ Filter \\ Size\end{tabular} & \begin{tabular}[c]{@{}c@{}}Flatten\\   Feature \\ Size\end{tabular} & \begin{tabular}[c]{@{}c@{}}CalTech\\   101\end{tabular} & \begin{tabular}[c]{@{}c@{}}CalTech\\   256\end{tabular} \\ \hline
 {\bf Num}  & {\bf Layer} & {\bf Filter Size} & { \bf Activation Shape} & {\bf Activation Size} &  {\bf CalTech-101} & {\bf CalTech-256} \\ \hline
1  & conv2d\_10   & 3x3x64& 35x35x96 & 117601 & 59.9\%& 24.0\%\\ \hline
2  & conv2d\_20 & 1x1x288& 35x35x64 & 78401& 52.7\%& 23.7\% \\ \hline
3  & conv2d\_30   & 3x3x96& 17x17x96  & 27745 & 71.3\%   & 32.0\%    \\ \hline
4  & conv2d\_40   & 1x1x768 & 17x17x192 & 55489 & 79.0\% & 44.1\% \\ \hline
5  & conv2d\_50   & 1x1x768 & 17x17x192 & 55489 & 79.7\% & 47.1\% \\ \hline
6  & conv2d\_60   & 1x1x768 & 17x17x192 & 55489 & 83.8\% & 52.2\% \\ \hline
7  & conv2d\_70   & 1x1x768 & 17x17x192 & 55489 & 81.1\% & 49.5\% \\ \hline
8  & conv2d\_80   & 3x1x384 & 8x8x384 & 24577 & 93.3\%& 74.2\%        \\ \hline
9  & conv2d\_90   & 1x1x2048& 8x8x448 & 28673& 85.3\% & 62.1\%\\ \hline
10 & conv2d\_94   & 1x1x2048& 8x8x192 & 12289& {\bf 94.4}\%& 78.2\%\\ \hline
11 & mixed10      & - & 8x8x2048 & 131073& 91.9\%& 77.9\%  \\ \hline
12 & avg-pool      & 8x8 [pooling] & 1x1x2048 &2049& 90.0\%& {\bf 79.1}\% \\ \hline
\end{tabular}
\label{table:layerwise}
\end{table*}

\begin{figure}[h]
    \centering
    \includegraphics[width=0.9\linewidth]{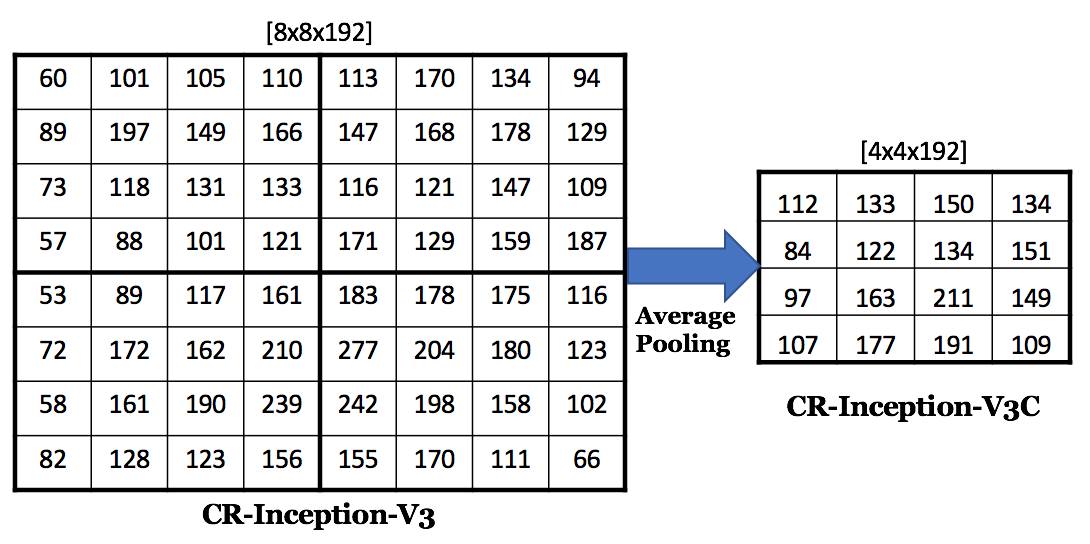}
    \caption{Average Pooling is applied; 8X8 sized 192 channels [8x8x192] are reduced to 4x4 sized 192 channels [4x4x192]}
    \label{fig:AvgPool}
\end{figure}

%\subsubsection{Results for Optimization}
Considering the feature size of the Layer~10 in Inception-V3, we evaluate by reducing the dimensions using standard reduction techniques, namely minimum pooling (MinPool), maximum pooling (MaxPool), and average pooling (AvgPool). The pooling in CBL was implemented on the [8x8,192] feature vector (Layer~10) with the filter size of 2x2 transforming into [4x4,192]. In CR-Inception-V3C,
we applied  AvgPool with a filter size [2x2] to the feature map of [8x8x192] extracted from Layer~10  and obtained the reduced feature map of [4x4x192]. 
As shown in Figure~\ref{fig:AvgPool}, the average pooling layer reduction is applied to the post-processing of CR-Generation. 
The same filter size was used for MaxPool and MinPool.

Table~\ref{FeatureAcc} shows the accuracy for  CR-Inception-V3C by using the filter size based on the three pooling techniques. Based on the analysis with all of the datasets, CR-Inception-AvgPool showed the accuracy drop with an average of 1\% in Top-1 accuracy when comparing with 12K or the original Layer~10 Accuracy. The interesting observation through this evaluation was the 12K CRL model's Top-5 Accuracy outperformed on all datasets compared with other available models.

\begin{table*}[h]
    \centering
    \renewcommand{\arraystretch}{1.2}
    \caption{Feature Pooling and Top  Accuracy\\12K: Original;    AVP: Average Pooling; MIP: Min Pooling; MAP: Max Pooling}
    \begin{tabular}{|c|c|c|c|c|c|c|c|c|c|c|c|c|c|c|c|c|}\hline
     	&  \multicolumn{4}{|c|}{\bf ImageNet-1K}	& \multicolumn{4}{|c|}{\bf CalTech-101} & \multicolumn{4}{|c|}{\bf CalTech-256} & \multicolumn{4}{|c|}{\bf CIFAR-100} \\\hline			
            &	12K	& AVP &	MIP	& MAP&		12K	& AVP &	MIP	& MAP&		12K	& AVP &	MIP	& MAP&		12K	& AVP &	MIP	& MAP \\\hline
Feature\#	& 12288	& 3072 & 3072 & 3072	 &	12288	&3072 & 3072 & 3072	 &	12288	& 3072 & 3072 & 3072	 &	12288	& 3072 & 3072 & 3072	  \\\hline\hline
{\bf Top 1} &	73.93 &	74.07 &	72.47	& 70.12	& 93.96	 &93.69	 & 93.46 &	91.89 &	 77.87 &	77.78	& 74.83	& 71.14 &	 57.96 &	57.63 &	55.76 &	55.25 \\\hline
{\bf Top 2} &	83.7 &	83.89 &	81.92&	79.73 &	97.23&	 97.23&	97.08& 	96.04 &	84.59 &	{\bf 84.6} &		81.75 &		78.31 &		70.61 &		70.38 &		68.73 &		67.95 \\\hline
{\bf Top 5} &		93.31 &		90.56 &		88.41 &		86.59 &		 99.15 &		98.73 &		98.42 &		97.89 &		 92.43 &		89.83 &		86.99 &		84.37 &		 90.5 &		83.42 &		81.89 &		81.48\\\hline
    \end{tabular}
    \vspace*{1mm}
    
%12K: Original;    AVP: Average Pooling; MIP: Min Pooling; MAP: Max Pooling
    \label{FeatureAcc}
\end{table*}

\subsubsection{Results from Data Imbalance Experiments}
Most of the deep learning suffered from the data imbalance problem.
Our experiments show that CRL is not sensitive to the data imbalance issue. 
Since CalTech-101 and CalTech-256 are imbalanced datasets, we have conducted experiments to show their classification performance is independent of the number of input. 

In order to analyze the imbalanced data issue in the CRL model, we have evaluated the CR Generation with a varying number of images and accuracy. As
Table~\ref{table:ImageCount} shows the CR Generation accuracy for the image sets of 20, 30, 60, 100 and all. All represented in the training dataset (i.e., 70\%) for any given class. The results show that the accuracy of CR-based classification does not vary significantly for a varying number of images. This effect is clearly shown with the imbalanced datasets such as CalTech-101 and CalTech-256.  For example, CalTech-101, Class {\it airplanes} have 560  images with the class accuracy of 97.9\% while {\it camera} has 35 images with the class accuracy of 96.8\%. This indicates that the image imbalance problem does not affect the classification accuracy in the CRL model.

\begin{table}[h]
\centering
\renewcommand{\arraystretch}{1.2}
\caption{Image Classification Accuracy\\ for an Increasing Image\#}
\begin{tabular}{|c|c|c|c|c|c|}\hline
Dataset &\multicolumn{5}{|c|}{Image Count \& Accuracy}\\ \cline{2-6}
           & 20   & 30   & 60 & 100 & All \\ \hline
ImageNet-1K & 69.5\%  & 70.9\% & 72.5\% & 73.0\% & 73.9\% \\ \hline
CalTech-101 & 91.9\% & 93.5\% &  93.6\%  & 93.8\%    & 93.9\%    \\ \hline
CalTech-256 & 74.4\% & 75.8\% & 77.3\% & 77.8\% & 77.9\% \\ \hline
CIFAR-100   & 50.9\% & 53.4\% & 55.9\% & 56.8\% & 57.9\% \\ \hline
\end{tabular}
\label{table:ImageCount}
\end{table}

\subsection{Model Performance Comparison}\label{ED-SOTA-Performance}
The CRL model's performance is evaluated by comparing with Inception-V3 pre-trained model retrained with the target domain. The CRL model's performance is calculated based on AFM Generation Time (refer to Section \ref{Section_CR_Definition}), CR-based Inference Time (refer to Section \ref{Section_CR_Inferencing}) and CR Model Generation Time (refer to Section \ref{Section_CR_Generation}). 

Table~\ref{table:Performace} shows the comparison of the CRL model's overall time vs. the time taken for retraining the dataset using the Inception-V3 pre-trained model. The pre-trained model was run on the same system specification as the CRL model. Pre-training of the Inception-V3 Model was stopped at a reported number of epochs as the time taken was significantly higher than that of the CRL model. The CRL model with three datasets, (CalTech-101, CalTech-256, and CIFAR-100) have an average of 99\% time reduction that is a significantly reduced time compared with that for the original Inception-V3 model. Within the same time window and based on the same pre-trained model, the Inception-V3 model performance has not reached the accuracy published in \cite{kornblith2018better}. The CRL model's overall time shows genuinely outstanding performances in the target domains, even if the models never learned from the target domain.  

\begin{table*}[hbt!]
\renewcommand{\arraystretch}{1.2}
    \centering
    \caption{Transfer Learning Performance Analysis: CRL vs. Inception-V3 \cite{Pretrain97:online}\\ Pretrained Model: Inception-V3 with ImageNet-1K}
    %(FET- Feature Extraction Time, CR Generation - Class Representative Generation, Pretrained - Pretrained InceptionV3 Model \cite{Pretrain97:online}] Learning Rate Pretrained Model is 0.1}
    \begin{tabular}{|c|c||c|c|c|c||c|c|c|}
    \hline
     \multicolumn{2}{|c||}{\bf Domain} & \multicolumn{4}{|c||}{\bf CRL} &  \multicolumn{3}{c|}{\bf Inception-V3 \cite{Pretrain97:online}} \\ \hline
   Source & Target & Step & Time  & Overall Time & Accuracy &   Time &  Epoch & Accuracy \\\hline
  \multirow{9}{*}{ImageNet-1K} &   \multirow{3}{*}{\bf CalTech-101} & AFM Generation & 5m 21s  & \multirow{3}{*}{7m 12s} & \multirow{3}{*}{94.40\%}  &   \multirow{3}{*}{4257m 55s}  &  \multirow{3}{*}{40 Epochs} & \multirow{3}{*}{88.73\%}\\\cline{3-4}
   &  & CR Model Generation & 1m 31s &  &  &  &  &   \\ \cline{3-4}
   &   & CR Inferencing & 20s &  & & & &    \\ \cline{2-8}\cline{2-9}
   &   \multirow{3}{*}{\bf CalTech-256} & AFM Generation & 10m 14s & \multirow{3}{*}{13m 53s}& \multirow{3}{*}{78.20\%}    & \multirow{3}{*}{14193m 58s} & \multirow{3}{*}{14 Epochs} & \multirow{3}{*}{59.26\%}\\\cline{3-4}
   &  & CR Model Generation & 1m 54s &  &  & &    &  \\ \cline{3-4}
  &   & CR Inferencing & 1m 45s &  &  & &   &  \\ \cline{2-8}\cline{2-9}
  &   \multirow{3}{*}{\bf CIFAR-100} & AFM Generation & 13m 12s & \multirow{3}{*}{15m 44s} & \multirow{3}{*}{57.96\%}   & \multirow{3}{*}{26941m 51s} & \multirow{3}{*}{10 Epochs} &   \multirow{3}{*}{50.30\%}\\ \cline{3-4}
  &   & CR Model Generation & 56s &  & &  &     &   \\ \cline{3-4}
  &   & CR Inferencing & 1m 36s &  &  & &   &  \\ \hline
    \end{tabular}
    \label{table:Performace}
    \vspace{1mm}
\end{table*}

\subsection{Comparison with Lightweight Classification Models}\label{ED-SOTA-Classification}

The two CRL models, namely CR-Inception-V3 [based on 12K Layer~10] (refer to Figure~\ref{fig:Activations_LayerWise}) and CR-Inception-V3C  [based on AvgPool] (refer to Figure~\ref{fig:AvgPool}) were considered for evaluation. The CRL models are compared with the state-of-the-art mobile Deep Learning models such as Mobile-Net-v1 \cite{howard2017mobilenets}, Mobile-Net-v2 \cite{sandler2018mobilenetv2} and NasNet-Mobile \cite{zoph2018learning}. The Inception-V3 Model accuracy is also compared  as the CRL model is based on Inception-V3 (refer to Section~\ref{ED-Accuray}). The  state-of-the-art accuracies shown in \ref{table:Pretrained} were based on the work by Kornblith et al.  \cite{kornblith2018better}. The another work by Kornblith et al. \cite{tensorflowSlim:online} was based on the performance of checkpoints from TensorFlow-Slim repository. 

The following a brief review of Light-Weight Classification models that are compared with CRL.
\begin{itemize}
    \item {\bf{MobileNet-v1}} MobileNet-v1 is one of pioneer work in light-weight convolution neural network. A efficient low latency model is achieved using Depthwise Convolution Filters \cite{howard2017mobilenets}
    \item {\bf{MobileNet-v2}} MobileNet-v2 is extention of MobileNet-v1, they improve light-weight model using inverted residual module with linear bottleneck. MobielNet-v2 1.4 is a version with neural network image input size of 224x224 and multipliers set to 1.4 \cite{sandler2018mobilenetv2}.
    \item {\bf{NASNet-A-Mobile}} NASNet is Convolution Neural Network where the Semantic Space is transfered from smaller dataset to bigger dataset using a Reinforcement Learning Search Method called Neural Architecture Search (NAS) \cite{zoph2018learning}.
\end{itemize}

The comparison between the mobile models and CRL is conducted in terms of the computational cost, model size, and accuracy. The computation cost is usually defined based on floating-point operations (FLOPs) and parameters. The FLOPs of the CRL model are less than the one for prediction with the base model, i.e., Inception-V3. As shown in Table~\ref{table:Pretrained}, the number of parameters for the CRL model is Nil as there is no traditional learning component in the CRL model. The CRL model's computational cost outperforms all the other models.

The CRL model's size is given based on the size of each CRL of class. The significant difference between the traditional deep learning model and the CRL model is that the model size is dependent on the number of classes rather than the number of layers or size of the layer. The size as listed in Table~\ref{table:Pretrained}, includes two parts the size of the pre-trained model plus the per CR per class size, i.e., 0.15MB for CR-Inception-V3 and 0.06MB for CR-Inception-V3C.

In Table \ref{table:Pretrained}, the two CRL models were compared with other mobile models. These models outperform the existing mobile models. Also, these models perform better than the original Inception model with CalTech-101 and CalTech-256 datasets. Their performances are reasonably comparable to the original ones with ImageNet-1K Dataset. However, the CRL models do not perform well on the CIFAR-100 Dataset. Overall, the CRL models are better than state-of-the-art mobile models if the target domains are similar to the source models. Otherwise, as seen from the CIFAR-100 model, the performance did not meet expectations. It is because there is a considerable gap between the source and target domains, and this gap may result from the lack of learning in the target domain.

This result confirms that the 
CRL model can be used to validate the distribution of data in terms of dissimilarities and similarities of CRs.  
The classification accuracy can be estimated based on the CR distribution model.  
Furthermore, outliers of data can be normalized, or mislabeled images can be detected with a CR.

%The optimal distribution,  MCDD (multi-class discrimitive distribution) \cite{mayanka} can be determined with the CRs.

\begin{table*}[hbt!]
    \renewcommand{\arraystretch}{1.2}
    \centering
      \caption{Comparison of Class Representative with Pre-trained Models(\cite{kornblith2018better}, \cite{Pretrain65:online})}
    \begin{tabular}{|c|c|c|c|c|c|c|c|c|}
        \hline
        & & & & & \multicolumn{4}{|c|}{Accuracy} \\ \cline{6-9}
         Model & Parameters\textsuperscript{a} & Feature Size & Image Size & Model Size &   ImageNet-1K& CalTech256 & CalTech101 & CIFAR100\\\hline
         MobileNet-V1 \cite{howard2017mobilenets} & 3.2M & 1024 &224& 16MB &70.6\% & N/A& 90.7\% & 70.9\% \\ \hline
         MobileNet-V2 \cite{sandler2018mobilenetv2} & 2.2M & 1280 &224& 13MB & 72\% & N/A & 91.26\% & 70.6\% \\ \hline
         MobileNet-V2 (1.4) \cite{sandler2018mobilenetv2}  & 4.3M & 1792 &224 & 24MB & 74.7\% & N/A & 91.83\% & 73.4\% \\ \hline
         NASNet-A Mobile \cite{zoph2018learning} & 4.2M & 1056 & 224 & 20MB & 74\% & N/A & 91.52\% & {\bf 73.6\%} \\ \hline
         Inception-V3 \cite{szegedy2016rethinking} & 21.8M	& 2048 & 299 & 89MB & 78.8\% & N/A & 92.98\% & 76.2\% \\ \hline
         CR-Inception-V3 (ours) & N/A & 12288 & 299 & 0.15MB/CR\textsuperscript{*}&73.93\% & {\bf 77.87\%} &{\bf 93.96\%} & 57.96\% \\ \hline
         CR-Inception-V3C (ours) & N/A & 3072 & 299& 0.06MB/CR\textsuperscript{*}& {\bf 74.07\%} & 77.78\% & 93.69\% & 57.63\% \\ \hline
    \end{tabular}
    \label{table:Pretrained}
        \vspace*{1mm}
        
    \footnotesize{\textsuperscript{*} CR-Inception-V3 uses Inception-V3 as its pre-trained model, which is 89MB.}
\end{table*}

\subsection{Comparison with Zero-Shot Learning Algorithms}\label{ED-SOTA-ZSL}
In this section, we evaluate the CRL model using three different evaluations; Recognition Task-based Accuracy, Accuracy with an increasing number of instances of the unseen dataset, and comparison with state-of-the-art Zero-Shot Learning (ZSL) approach. For this section, we consider two versions of the CRL model; (i) Inception-V3 based and (ii) VGG-19 based model. In Tables~\ref{Table:RecognitionTask} and \ref{tab:IncreasingIns}, the performance of Inception-V3 model-based CRL  (CR-Inception-V3) in the ZSL perspectives was presented. 

The CRL model is capable of recognizing the target labels (unseen data) without having the source labels (seen data) as an option. This shows the advantage and ability as a classification model (see Section~\ref{ED-SOTA-Classification}). Table~\ref{Table:RecognitionTask} shows two versions of the recognition tasks with testing data from target set ($\mathbb{T}$); $\mathbb{T}\Rightarrow\mathbb{T}$ when the testing label could be only from the target set $y^* \in \mathbb{T}$  and $\mathbb{T}\Rightarrow\mathbb{S}\cup\mathbb{T}$ when the testing label could be from both the source set and target set $y^* \in \mathbb{S} \cup \mathbb{T}$. For this experiment, we consider all instances (70\% of the dataset) from the dataset to generate CRs. The increase in the number of labels in the dataset was compared with the accuracy. There were significant drops when the source set was also considered. The interesting observation is that the Heterogeneous Domain (HD) such as CIFAR-100 does not have a significant drop in accuracy, which makes sense, as CIFAR-100's CR space does not overlap with ImageNet-1K's CR space (see Figure~\ref{fig:TSNE}).

\begin{table}[h]
\renewcommand{\arraystretch}{1.2}
\centering
\caption{Accuracy for CR-Inception-V3 \\ Zero-Shot Learning Tasks}
\begin{tabular}{|c|c|c|c|}
\hline
\multirow{2}{*}{Dataset} & \multicolumn{3}{c|}{Recognition Task Accuracy} \\ \cline{2-4}
 & Accuracy & $\mathbb{T} \Rightarrow \mathbb{T}$  & $\mathbb{T} \Rightarrow \mathbb{S} \cup \mathbb{T}$  \\ \hline
\multirow{2}{*}{CalTech-101} & Top 1 & 93.9\% & 87.4\% \\ \cline{2-4} 
 & Top 5 & 99.1\% & 97.5\% \\ \hline
\multirow{2}{*}{CalTech-256} & Top 1 & 77.8\% & 40.9\% \\ \cline{2-4} 
 & Top 5 & 92.4\% & 54.9\% \\ \hline
\multirow{2}{*}{CIFAR-100} & Top 1 & 57.9\% & 57.4\%\\ \cline{2-4} 
 & Top 5 & 90.5\% & 83.4\% \\ \hline
\end{tabular}
\label{Table:RecognitionTask}
\end{table}

Table~\ref{tab:IncreasingIns} shows the performance of CR-Inception-V3 with one image from each class to ten images from each class. For this experiment, we considered only ($\mathbb{T}\Rightarrow\mathbb{T}$) setting. The interesting to see that the CRL model with just ten images from each class all the dataset Top-1 accuracy reach more than 75\% of accuracy achieved when all instances are used.

\begin{table}[h]
\renewcommand{\arraystretch}{1.2}
\centering
\caption{CR-Inception-V3 Accuracy with \\ Increasing Target Instances (\#Ins.)}
\begin{tabular}{|c|c|c||c|c||c|c||c|c|}
\hline
\multirow{2}{*}{\textbf{\#Ins.}} & \multicolumn{2}{|c||}{\textbf{CalTech-101}} & \multicolumn{2}{|c||}{\textbf{CalTech-256}} & \multicolumn{2}{|c||}{\textbf{CIFAR-100}} & \multicolumn{2}{|c|}{\textbf{ImageNet-1K}} \\ \cline{2-9}
 & T-1 & T-5 & T-1 & T-5 & T-1 & T-5 & T-1 & T-5 \\\hline
\textbf{1} & 70.2 & 83.4 & 41.4 &54.8 & 20 &38 & 32.5 & 50.1 \\ \hline
\textbf{2} & 79.2 & 92.3 & 51.7 & 66.7 & 25.5 & 52.2 & 44 & 64.9 \\ \hline
\textbf{3} & 85.3 & 95.2 & 56.5 & 72.1 & 30.3 & 56.9 & 50.9 & 72.3 \\ \hline
\textbf{4} & 85.6 & 96.7 & 61.5 & 77.1 & 36 & 62.9 & 54.9 & 76.2 \\ \hline
\textbf{5} & 86.8 & 97 & 63.6 & 78 & 38 & 66.7 & 58.1 & 79 \\ \hline
\textbf{6} & 87 & 97.6 & 65.1 & 79.4 & 40.9 & 69.5 & 60.3 & 81.2 \\ \hline
\textbf{7} & 89.4 & 97.7 & 67.2 & 82.2 & 42.4 & 70.8 & 61.3 & 82.2 \\ \hline
\textbf{8} & 90.7 & 98 & 68.2 & 82.4 & 44.6 & 73.6 & 63.4 & 83.6 \\ \hline
\textbf{9} & 90.8 & 97.9 & 69.8 & 83.8 & 44.8 & 74.1 & 64.6 & 84.6 \\ \hline
\textbf{10} & 91.2 & 98.2 & 70 & 84.4 & 45.8 & 75.4 & 65.4 & 85.2 \\ \hline
\textbf{all} & 93.9 & 99.1 & 77.8 & 92.4 & 57.9 & 90.5 & 73.9 & 93.3 \\ \hline
\end{tabular}
\label{tab:IncreasingIns}
\end{table}

Table~\ref{tab:zsl} shows the comparison of state-of-the-art Zero-Shot Learning methods briefly introduced in Table~\ref{tab:relatedWorkTable}. 
The following is a brief review of the ZSL models, which are compared with CRL.
\begin{itemize}
    \item \textbf{Deep WMM-Voc}: Deep Weight Maximum Margin Vocabulary (Deep WMM-Voc) is Classifier-based ZSL, which uses a joint embedding for visual features and semantic words. The results generated for this comparison done based on the same settings proposed in this work \cite{fu2019vocabulary,fu2016semi}.
    \item \textbf{SAE}: Semantic Auto-Encoder (SAE) is an encoder-decoder paradigm that projects visual features into semantic representation spaces such as Attribute Space and use the decoder to reconstruct the original visual features \cite{kodirov2017semantic}.
    \item \textbf{ESZSL \& Deep-SVR}: ESZSL and Deep-SVR are an Attribute Space-based ZSL model. ESZSL implements a linear model together with regularizers for attribute learning, whereas Deep-SVR uses a probabilistic attribute classifier \cite{romera2015embarrassingly,lampert2013attribute}. 
    \item \textbf{Embed}: Embed is Instance-Based ZSL using Label Embedding Space. Embed uses Multi-modal fusion semantic representation with RNN \cite{zhang2017learning}.
    \item \textbf{DeViSE \& ConSE}: DeViSE and ConSE are Label Embedding Space-based ZSL Models. Both benefit from Convolution Neural Networks, where DeViSE uses SoftMax Classifier, and ConSE uses a ranking algorithm for prediction \cite{norouzi2013zero,frome2013devise}.
    \item \textbf{AMP}: AMP is a Label Embedding Space-based ZSL model. Absorbing Markov Chain Process is formulated on a semantic graph that provides processing efficiency in ZSL \cite{Fu_2015_CVPR}.
\end{itemize}

Table~\ref{tab:zsl} shows that  the performance of the CRL model is superior in both cases, with 3000 instances and all instances. In the case of the 3000 instances, the CRL model's Top-1 has a 27\% increase from the top performer, Deep WMM-Voc. In the case of all instances, the CRL model's Top-1 is significantly higher than others;
on average, the CRL model's Top-1 accuracy is 3 times higher than Deep WMM-Voc's.

\begin{table}[h]
    \renewcommand{\arraystretch}{1.2}
    \centering
    \caption{Comparison between the CRL model with VGG-19/ImageNet-1K and State-of-the-art ZSL Models}
    \begin{tabular}{|c|c|c||c|c|}
        \hline
         \multirow{2}{*}{Methods}& \multicolumn{2}{|c||}{\bf 3000 Instances}&\multicolumn{2}{|c|}{\bf All Instances}  \\ \cline{2-5}
         & Top-1 & Top-5 & Top-1 & Top-5 \\ \hline
         CRL with VGG-19 (ours) & {\bf 11.78} & {\bf 25.52} & {\bf 31.6} & {\bf 55.1} \\ \hline
         Deep WMM-Voc \cite{fu2019vocabulary,fu2016semi}& 9.26 & 21.99 & 10.29 & 23.12 \\ \hline
         SAE \cite{kodirov2017semantic} & 5.11 & 12.26 & 9.32 & 21.04 \\ \hline
         ESZSL \cite{romera2015embarrassingly} & 5.86 & 13.71& 8.3 & 18.2 \\ \hline
         Deep-SVR \cite{lampert2013attribute} &5.29&13.32& 5.7 & 14.12 \\ \hline
         Embed \cite{zhang2017learning}&-&-& 11.00 & 25.7 \\ \hline
         ConSE \cite{norouzi2013zero}&5.5&13.1& 7.8& 15.5 \\ \hline
         DeViSE \cite{frome2013devise} &3.7&11.8& 5.2& 12.8 \\ \hline
         AMP \cite{Fu_2015_CVPR}& 3.5 &10.5& 6.1& 13.1 \\ \hline
         %Chance&2.78\it{e}-3&-&- \\ \hline
    \end{tabular}
    \vspace{1mm}
    
    \footnotesize{*Results reported for SOTA Models are from Fu et al.\cite{fu2019vocabulary}. 
   The CRL model is configured with the same settings such as VGG-19 with 3000 instances i.e., 3 images per class and all 50000 instances i.e., 50 images per class}
    \label{tab:zsl}
\end{table}

\section{Discussion}

The limitations of the CRL model are that  
a source environment (except the fully connected layers)  is still required for generating 
a feature map for any input in the testing. 
The size of the source environment  may be too big to fit in low-end devices like mobile devices.
In our research, we can provide a cloud service for the feature map generation as a basic interface, 
such as Application Programming Interfaces (APIs) for lightweight mobile applications.

The Class Representative (CR) generation was obtained by extracting the abstraction of the distribution 
of each feature in the Class Representative Feature Space (CRFS) of an input image. For the purpose, we used a simple average mean approach.
Thus, a CR can be sensitive to outliers and sample size bias of the CRFS. The CRL model was extremely strong 
at the Top-5 inferencing compared to Top-1 inferencing (see Table~\ref{FeatureAcc}).
The CR computation might not be accurate due to  bias or unexpected outliers.
This indicates that the high similarity between some CRs can lead to  misclassification. 
To overcome the limitation of the CR Generation, We will explore an advanced optical model such as the Fisher Vector and Gaussian-Mixture-Model.
We also can use  unsupervised deep learning techniques such as the autoencoder  
in learning efficient data codings to reduce the CR's Feature Space to a more optimal representation.
We can further extend it to determine the common and unique features of the CR vectors 
and find the weights that maximize the uniqueness between CRs.

\section{Conclusion}
We presented the Class Representative Learning (CRL) model that is based on class-level classifiers, built class-by-class, that would be a representative of instances of a specific class by utilizing activation features of Convolutional Neural Networks responding to the new cases.
The characteristics of the CRL are high efficiency, being compact and lightweight.  It was possible because the CRs can be generated in a parallel and distributed manner, and the inferencing can be conducted through matching new inputs with CRs. 
Comprehensive evaluations have been conducted with the CRL model, 
compared to the state-of-the-art approaches both in classification and zero-shot learning
using the four benchmark datasets. 
The CRL model was shown to increase accuracy and reduce considerable times in building CNN based classifiers.

\section{Acknowledgment}

This work was partially supported by NSF CNS \#1747751 and CNS \#1650549.
%The authors would like to thank

\balance

% \bibliographystyle{IEEEtran}
% \bibliography{references}
%\renewcommand*{\bibfont}{\small}
%\printbibliography
\end{document}